\PassOptionsToPackage{table,dvipsnames}{xcolor}
\documentclass{article}
\usepackage[final]{colm2026_conference}

\usepackage{microtype}
\usepackage{hyperref}
\usepackage{url}
\usepackage{booktabs}
\usepackage{graphicx}
\usepackage{xcolor}
\usepackage{longtable}
\usepackage{enumitem}
\usepackage{makecell}
\usepackage{multirow}
\usepackage[most]{tcolorbox}
\usepackage{xspace}
\usepackage{caption}
\usepackage{wrapfig}
\usepackage{placeins}

\usepackage{lineno}

\definecolor{darkblue}{rgb}{0, 0, 0.5}
\definecolor{ForestGreen}{rgb}{0.133, 0.545, 0.133}
\hypersetup{colorlinks=true, citecolor=darkblue, linkcolor=darkblue, urlcolor=darkblue}

\usepackage{amsmath,amsfonts,bm}

\def\eqref#1{equation~\ref{#1}}

\def\1{\bm{1}}

\DeclareMathAlphabet{\mathsfit}{\encodingdefault}{\sfdefault}{m}{sl}
\SetMathAlphabet{\mathsfit}{bold}{\encodingdefault}{\sfdefault}{bx}{n}

\newcommand{\modelname}{\textsc{ASI-Arch}\xspace}

\title{Neural Architecture Discovery via Autonomous Evolution}

\author{
\begin{tabular}{c}
Weixian Xu\textsuperscript{1,2}\thanks{Co-first authors.} \quad
Yixiu Liu\textsuperscript{2,3,4}\footnotemark[1] \quad
Yang Nan\textsuperscript{2}\footnotemark[1] \\
Lyumanshan Ye\textsuperscript{2,4} \quad
Xiangkun Hu\textsuperscript{2,3} \quad
Zhen Qin\textsuperscript{5} \quad
Pengfei Liu\textsuperscript{2,3,4}\thanks{Corresponding author: \href{mailto:pengfei@sjtu.edu.cn}{\texttt{pengfei@sjtu.edu.cn}}} \\[0.5em]
\small\normalfont\textsuperscript{1}Zhiyuan College, Shanghai Jiao Tong University \quad
\textsuperscript{2}Generative AI Research Lab (GAIR) \\
\small\normalfont\textsuperscript{3}Shanghai Innovation Institute \quad
\textsuperscript{4}Shanghai Jiao Tong University \quad
\textsuperscript{5}TapTap
\end{tabular}
}

\begin{document}

\ifcolmsubmission
\linenumbers
\fi

\maketitle

\begin{abstract}

Recent progress in LLM agents has advanced the prospect of autonomous research. Yet whether AI can complete difficult long-horizon tasks, especially those that advance AI research itself, remains largely unexplored. We present \modelname, a system for \textbf{AI-driven AI research} that autonomously conducts neural architecture research through a closed-loop \emph{research-experiment-analyze-update} process. Applied to linear attention, \modelname ran 1,773 iterative experiments and discovered 105 state-of-the-art architectures. Its best architecture improves over DeltaNet by nearly three times the gain achieved by Mamba2. Beyond the final performance gains, we analyze the contributions of different parts of the framework in this hard research setting, shedding light on what enables autonomous progress in complex AI research tasks.

\end{abstract}

\section{Introduction}

As AI systems continue to improve, the dominant research paradigm is shifting from human-AI collaboration toward increasingly autonomous scientific discovery, giving rise to the vision of an ``AI Scientist'' capable of conducting research with limited human oversight~\citep{Boiko2023Autonomous, chen2025ai4researchsurveyartificialintelligence}. A central next question is whether AI can in turn accelerate progress in AI itself, especially on complex and technically demanding research problems. We therefore study \textbf{AI-driven AI research}: AI systems that autonomously investigate and advance AI itself---generating hypotheses, running experiments, and synthesizing findings with minimal human intervention.

Neural architecture discovery is among the most challenging and impactful domains for this vision. Each major leap---from CNNs~\citep{lecun1995convolutional} to Transformers~\citep{vaswani2017attention}---has come from architectural innovation. Moving beyond Neural Architecture Search (NAS), which is limited to human-defined spaces~\citep{zoph2016neural, elsken2019neural}, we present \modelname: a paradigm shift from \emph{automated optimization} to \emph{automated innovation}. Leveraging modern LLMs~\citep{brown2020language, openai2023gpt4}, \modelname autonomously hypothesizes novel architectural concepts, implements them as code, and validates performance through rigorous experimentation~\citep{chen2023llmatic, cheng2025language, chen2025ai4researchsurveyartificialintelligence}. Over 1,773 experiments and approximately 20,000 total GPU hours on linear attention~\citep{katharopoulos2020transformers, choromanski2020rethinking}, it continuously discovered 105 SOTA architectures, with the largest relative improvement over DeltaNet reaching three times that achieved by Mamba2. We further perform \textbf{cross-family transfer} to Mamba2, GLA~\citep{yang2024gla}, and HGRN2 to test whether AI-discovered design principles generalize beyond the original architecture family.

In this work, we make the following contributions: \textbf{(1)} We show that an AI system can autonomously complete the full cycle of complex neural architecture design; \textbf{(2)} We open-source the complete \modelname framework\footnote{\raisebox{-0.12em}{\includegraphics[height=0.95em]{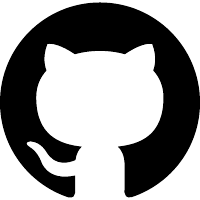}}\ \href{https://github.com/GAIR-NLP/ASI-Arch}{\nolinkurl{github.com/GAIR-NLP/ASI-Arch}}}, all 105 discovered SOTA linear attention architectures, and the entire experiment framework; \textbf{(3)} We show that AI-discovered architectural knowledge exhibits meaningful generality, transferring across model families and yielding consistent gains on Mamba2, GLA, and HGRN2.

\begin{figure*}[!t]
  \centering
  \includegraphics[width=\linewidth]{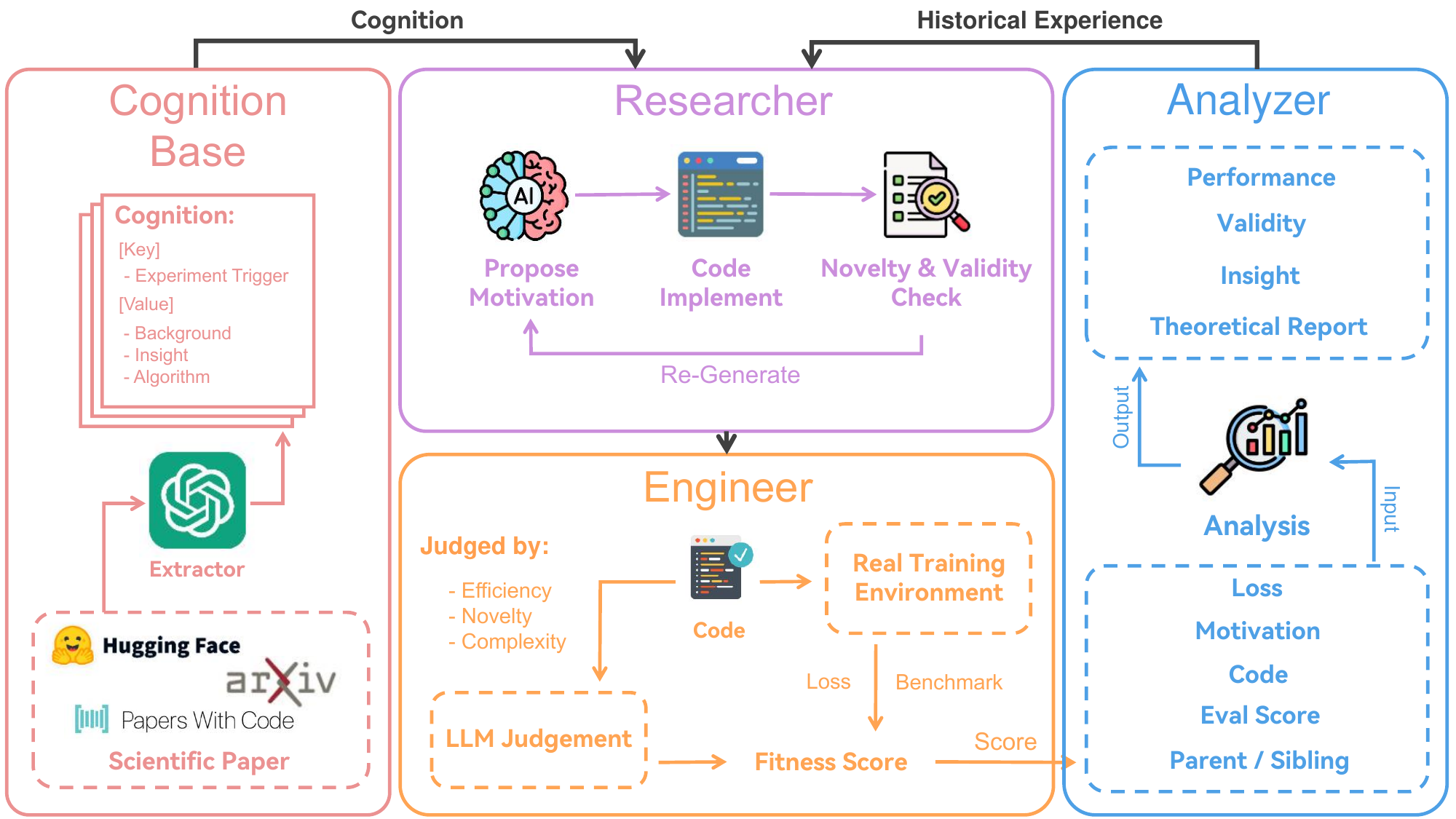}
  \caption{Overview of the \modelname framework. The Researcher proposes new architectures and implements their code using knowledge retrieved from the Cognition Base and experience accumulated from previous iterations. The Engineer then trains and evaluates the implementations, invoking a debugging agent when needed. The Analyzer interprets the resulting multifaceted experimental evidence and distills it into insights that guide the Researcher in the next iteration. This feedback loop supports iterative improvement.}
  \label{fig:method}
  \vspace{-8pt}
\end{figure*}

\section{Background}

\paragraph{AI scientific research.}
The use of AI as an autonomous scientist has grown rapidly alongside the capabilities of modern language models. A prominent class of such systems follows an evolutionary paradigm: given a research target, the AI iteratively proposes and evaluates designs in a closed loop, as exemplified by AlphaEvolve~\citep{novikov2025alphaevolve}, OpenEvolve~\citep{openevolve2025}, ShinkaEvolve~\citep{shinkaevolve2025}, GEPA~\citep{gepa2025}, and AdaEvolve~\citep{adaevolve2026}. A particularly compelling direction within this landscape is AI that studies AI itself---enabling systems to self-improve and self-iterate~\citep{Zhang2025DarwinGM, Schmidhuber2006Godel}, opening the possibility of systems that autonomously refine their own capabilities without much human intervention. Earlier work in this vein took the form of Neural Architecture Search (NAS)~\citep{zoph2016neural, elsken2019neural}, where the AI operates within a human-defined search space as a selection algorithm rather than a creative agent. \modelname adopts the same evolutionary framework and removes the fixed search space: rather than selecting from predefined candidates, the system freely hypothesizes novel architectures and navigates a far larger and more complex design landscape.

\paragraph{Efficient architectures.}
The Transformer architecture has dominated sequence modeling since its introduction, but its quadratic attention complexity has motivated extensive research into sub-quadratic alternatives~\citep{vaswani2017attention}. Three principal families with linear time complexity have emerged: Linear Attention, which linearizes the attention kernel via feature maps~\citep{katharopoulos2020transformers, choromanski2020rethinking, qin2022cosformer}; State-Space Models (SSMs) such as Mamba, which employ structured state-transition matrices~\citep{gu2023mamba, dao2024transformers}; and Linear RNNs such as RWKV and HGRN2, which use recurrent formulations with efficient linear-time inference~\citep{peng2023rwkv, qin2024hgrn2}. More recently, hybrid architectures like Jamba interleave multiple families to combine their respective strengths~\citep{lieber2024jamba}. This evolution has transformed the architectural landscape from a single dominant paradigm into a vast combinatorial space where optimal designs are highly task- and constraint-dependent. Manual exploration of this space is slow and expensive, typically requiring months of expert effort to produce a single state-of-the-art architecture. \modelname addresses this bottleneck by systematically exploring the space through automated multi-agent collaboration, making the automated discovery of novel architectures possible.

\section{Methodology}

As illustrated in Figure~\ref{fig:method}, \modelname is a closed-loop system for autonomous neural architecture discovery. Architecture design poses three compounding challenges that make it uniquely hard for autonomous AI: \textbf{(1)} LLMs have limited built-in knowledge of architecture design---the relevant literature is sparse and specialized; \textbf{(2)} implementing a model layer means modifying a large, intricate codebase where a single constraint violation silently corrupts training; \textbf{(3)} quality can only be measured by pre-training from scratch and running benchmarks---a costly, indirect chain vulnerable to reward hacking. These challenges directly shape the design of \modelname: each component is introduced to address one or more of them. The system consists of four modules: \textbf{Researcher}, \textbf{Engineer}, \textbf{Analyzer}, and \textbf{Experience Repository}. Together, they operate in a continuous loop augmented by an exploration-then-verification strategy.

\subsection{Researcher: propose new architecture}
\label{sec:evolution}

The Researcher proposes novel architectures by drawing on experimental history and human domain knowledge.

\paragraph{Model design and integrated implementation.}
For each cycle, the Researcher receives summaries of past architectures as context, then jointly articulates the architectural motivation and implements the corresponding code in a single pass. This integrated design-then-implement approach avoids the ``implementation drift'' common in two-stage pipelines, where an implementer working from a high-level description alone produces code that diverges from the intended design.

\paragraph{Novelty and sanity check.}
Before a proposal enters the expensive training pipeline, the Researcher gates it through a two-stage pre-training check. \emph{Novelty check}: the proposal's motivation is embedded and matched against the top-5 most similar historical motivations; a specialized LLM then judges whether the design represents genuine innovation or a superficial variant, filtering out redundant explorations. \emph{Sanity check}: a combination of automated code analysis and LLM-based review checks code quality, complexity, causal masking, and related implementation issues. Proposals that fail either check are rejected with specific feedback for revision. Together, these checks markedly improve the success rate of training runs and eliminate wasted GPU hours on fundamentally flawed submissions.

\subsection{Engineer: train and evaluate models}
\label{sec:validation}

The Engineer executes training and constructs the final fitness score.

\paragraph{Training execution.}
The Engineer runs each proposed architecture in a real interactive code environment. When a training run fails due to an implementation error, the full error log is returned to a dedicated debugging agent, which revises the code based on the concrete failure signal and resubmits the run. Beyond this self-correction loop, an automated monitor runs in parallel and immediately terminates any run that exhibits pathological signals: abnormally low loss, excessive compilation time, or training speed more than $2\times$ slower than baseline. Together these mechanisms maximize the fraction of GPU hours spent on genuinely viable candidates.

\paragraph{Fitness evaluation.}
To rank candidates reliably, we define a composite fitness that integrates quantitative and qualitative signals:
\begin{equation}
\text{Fitness} = \frac{1}{3}\left[\sigma(\Delta_{\text{loss}}) + \sigma(\Delta_{\text{benchmark}}) + \text{LLM}_{\text{judge}}\right]
\end{equation}
where $\sigma(\Delta_{\text{loss}})$ and $\sigma(\Delta_{\text{benchmark}})$ are sigmoid-transformed improvements over baseline (focusing on differences within 10\% to suppress noise). The sigmoid caps extreme values so that no single metric dominates, while still amplifying small but meaningful gains. Alongside these quantitative scores, we employ an LLM-as-judge ($\text{LLM}_{\text{judge}} \in [0,1]$) to capture dimensions that benchmark numbers alone cannot reflect---architectural complexity, computational efficiency, implementation elegance, and novelty relative to the baseline---providing a more comprehensive picture of each design's true merit.

\subsection{Analyzer: mine experimental insights}
\label{analysis}

After each experiment, the Analyzer converts the complete experimental record---performance metrics, training curves, and comparisons to parent and sibling nodes in the architectural phylogenetic tree---into a structured analysis that is archived in the Experience Repository. By comparing structurally similar architectures that differ in only a few components, the Analyzer can infer the isolated contribution of individual design choices, approximating an ablation study without requiring additional controlled runs. This accumulated analytical knowledge enables the system to improve not just by reusing prior designs, but by understanding \emph{why} certain designs succeed---generating insights that go beyond what any single paper in the literature can provide.

\subsection{Experience repository: shared memory across the search}
\label{sec:repo}

The Experience Repository is the shared memory that connects all modules and persists knowledge across the entire search.

\paragraph{Evolutionary candidate pool.}
All completed experiments are logged persistently. The system maintains a dynamically updated pool of the top-50 highest-scoring architectures. For each evolution step, the Researcher samples one parent from the top-10 for direct modification and four references from positions 11--50 for diverse examples. Updates to this pool are batched so that all parallel agents are more likely to work from a stable reference set. In addition, to ensure broader exploration diversity in the early stage of the search, we impose an initial cold-start period of 200 explorations during which the pool is frozen before convergence pressure sets in.

\paragraph{Cognition base.}
We distill approximately 100 seminal linear-attention papers into a structured knowledge base of roughly 150 \emph{cognition entries}. Each entry captures: (i) the \emph{applicable scenario}, (ii) the \emph{core algorithm}, and (iii) the \emph{historical context}. After each experiment, the Analyzer's summary of observed shortcomings is used as a retrieval query against the scenario field via embedding-based search; the top-3 matching cognitions are injected into the Researcher context for the next cycle. This retrieval-augmented mechanism gives the system access to targeted domain expertise on demand, substantially enriching its proposals without flooding the context with irrelevant background.

\subsection{Exploration-then-verification strategy}

We adopt a two-stage exploration-then-verification strategy to balance search breadth against evaluation rigor while also testing scalability. In the \textbf{exploration stage}, candidates are trained at 20M parameters on 1B tokens---fast enough to run thousands of experiments---and evaluated on 500 samples per benchmark. Only architectures that surpass DeltaNet on both quantitative and qualitative metrics advance. In the \textbf{verification stage}, survivors are scaled to 340M on 1B tokens with a strict causality check that verifies that perturbing later tokens does not change earlier-token hidden states, and the top candidates are further scaled to 1.3B on 15B tokens with six additional out-of-distribution benchmarks. Starting at small scale lets us screen a large candidate pool cheaply; the multi-scale verification then confirms that discovered gains persist as models grow, ruling out artifacts of the resource-constrained exploration setting.

\section{Main results}

We use DeltaNet~\citep{yang2024parallelizing} as the primary baseline throughout. Full experimental configuration details---model architecture, training protocol, dataset, and evaluation benchmarks---are provided in Appendix~\ref{sec:mid-eval}; additional reproducibility and cognition-source results are reported in Appendix~\ref{sec:additional-results}.

\paragraph{Search process.}

\begin{wrapfigure}{r}{0.48\linewidth}
\vspace{-8pt}
\begin{minipage}{0.48\textwidth}
  \centering
  \includegraphics[width=\linewidth]{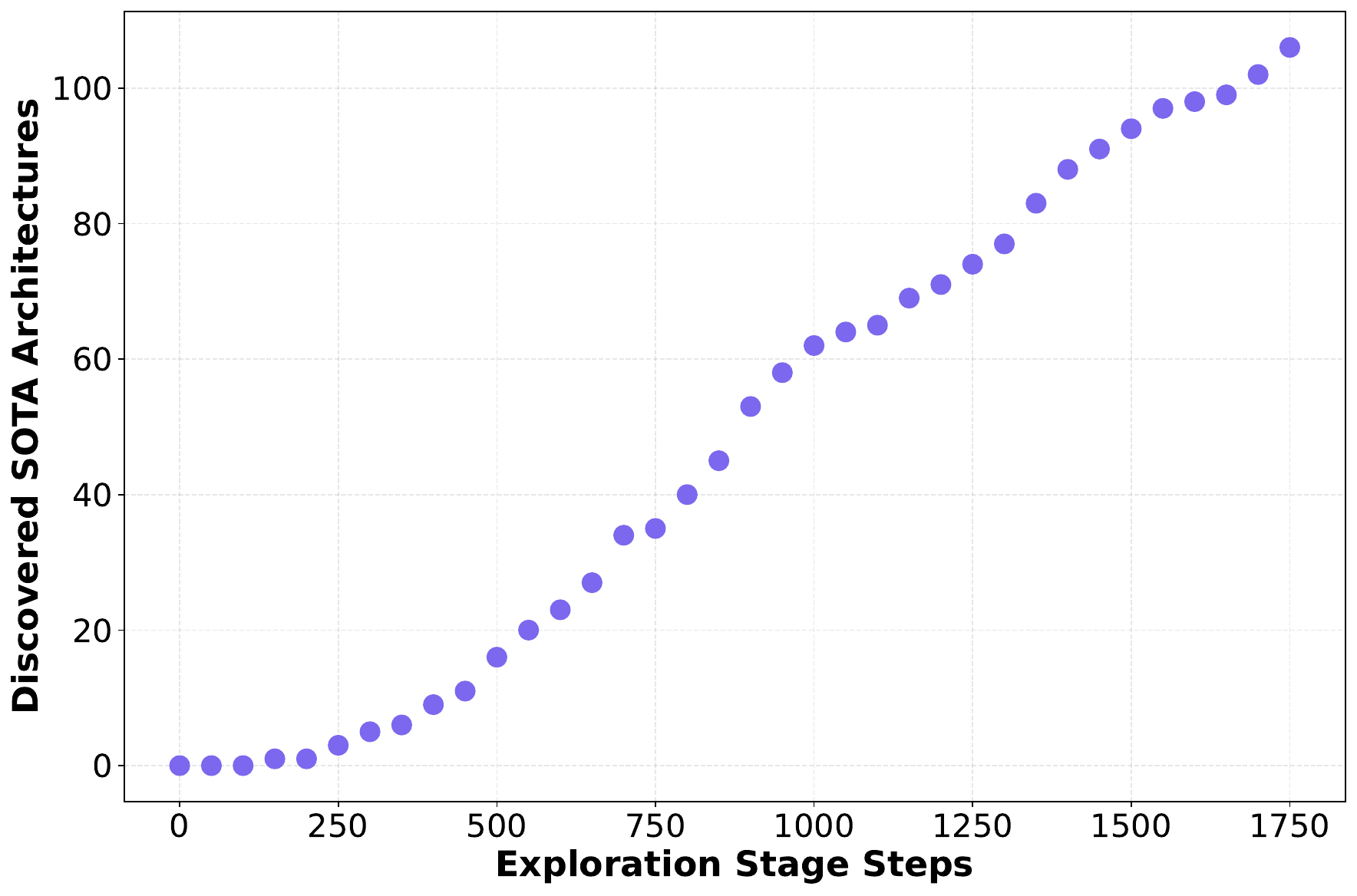}
  \caption{Cumulative count of discovered SOTA architectures vs.\ number of exploration experiments. The near-linear trend confirms that breakthroughs grow steadily with compute.}
  \label{fig:scaling}
\end{minipage}
\vspace{-10pt}
\end{wrapfigure}

The exploration stage ran 1,773 experiments at 20M scale, consuming approximately 10,000 GPU hours. Among these, 1,350 candidates outperformed DeltaNet on both quantitative and qualitative metrics and passed the exploration-stage filters. We selected approximately 150 of these for 340M-scale verification, of which \textbf{105 architectures achieved state-of-the-art results under our controlled 340M evaluation}. As shown in Figure~\ref{fig:scaling}, the cumulative count of SOTA discoveries grows at a near-linear rate throughout---allocating more compute translates reliably into more discoveries (see Section~\ref{sec:continuous-discovery} for a mechanistic explanation). We selected the top 5 architectures for large-scale validation.

\paragraph{Large-scale validation and discovered architectures.}
The five selected architectures were trained at 1.3B parameters on 15B tokens and evaluated on 16 benchmarks: 10 development benchmarks used throughout exploration (Wiki PPL, LAMBADA ppl/acc, PIQA, HellaSwag, WinoGrande, ARC-Easy/Challenge, Social-IQA, BoolQ) and 6 new generalization benchmarks (RACE, BBQ, MetaBench, QA4MRE-2011, SciQ, SWAG). Results are reported in Table~\ref{tab:id_ood_1p3b}.

\begin{wrapfigure}{r}{0.28\linewidth}
\vspace{-8pt}
\makebox[0pt][l]{%
\begin{minipage}{0.29\columnwidth}
  \centering
  \includegraphics[width=\linewidth]{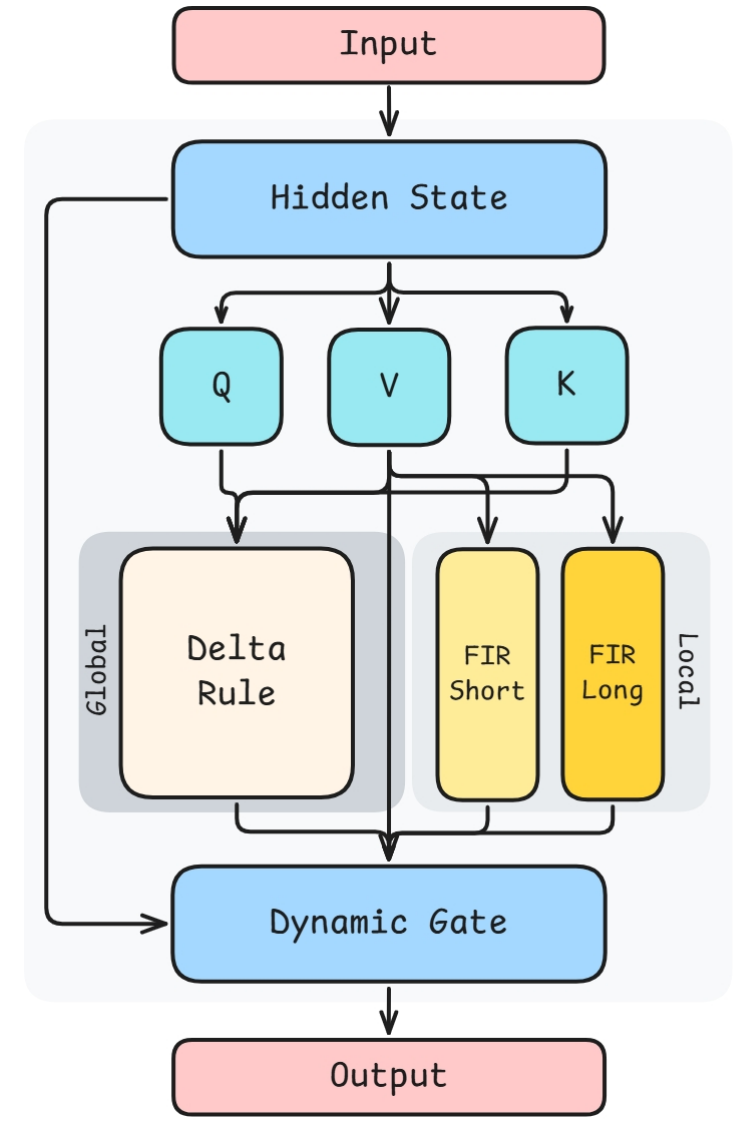}
  \caption{\textbf{HierGateNet}}
  \label{fig:hier_arch}
\end{minipage}}
\vspace{-12pt}
\end{wrapfigure}

Our best model improves over DeltaNet by \textbf{+0.97 points} on the overall average, nearly \textbf{3$\times$} the gain of the current strongest human-designed competitor, Mamba2 (\textbf{+0.34 points}). This improvement is achieved autonomously, without human architectural intuition. Across all five models, the average improvement over DeltaNet is +0.77 points on development benchmarks and +0.29 points on the new generalization set, confirming that gains are not artifacts of overfitting to the exploration benchmarks.

A common pattern among the strongest architectures is the use of multiple auxiliary paths coordinated through adaptive routing or gating. \textbf{PathGateFusionNet} uses a two-stage router that first separates direct copying from contextual processing, then distributes the contextual budget across short-range, long-range, and Delta-rule branches, with an additional lightweight residual path and head-wise output gate. \textbf{ContentSharpRouter} keeps a flat four-way router, but makes it content-adaptive and equips each head with a learnable temperature so routing can become sharper or softer depending on the input. \textbf{FusionGatedFIRNet} removes the softmax competition entirely and replaces it with independent sigmoid gates, allowing local and global paths to remain active at the same time; it also adds per-head retention to adjust the memory horizon of the Delta-rule branch.

\newpage
\textbf{HierGateNet}, illustrated in Figure~\ref{fig:hier_arch}, also adopts hierarchical routing, but factorizes it into a coarse local-versus-global choice followed by within-group selection, and introduces learnable path floors to prevent important routes from collapsing. \textbf{AdaMultiPathGateNet} pushes routing to the finest granularity with a unified gate that combines token-level, head-level, and global logits, stabilized by an epsilon floor and persistent entropy regularization.
Taken together, these variants indicate that the strongest designs consistently improve DeltaNet by making routing and gating more adaptive across paths and scales.

\begin{table*}[!ht]
    \centering
    \small
  \renewcommand{\arraystretch}{1.3}
  \setlength{\tabcolsep}{6pt}
  \begin{tabular*}{\linewidth}{@{\hspace{6pt}\extracolsep{\fill}}l|ccc|ccccc@{\hspace{6pt}}}
      \toprule
      \textbf{Benchmarks} & \textbf{DeltaNet} & \textbf{\makecell{Gated-\\DeltaNet}} & \textbf{Mamba2} & \textbf{PG} & \textbf{C} & \textbf{FG} & \textbf{H} & \textbf{AM} \\
      \midrule
      \rowcolor{gray!10}\multicolumn{9}{c}{\emph{ Development }} \\
      Wiki ppl$\downarrow$ & 17.00 & 16.84 & 16.66 & 16.18 & \underline{16.05} & \textbf{15.77} & 16.65 & 16.26 \\
     LMB ppl$\downarrow$ & 13.63 & 13.31 & 13.33 & \underline{12.62} & 13.45 & \textbf{12.34} & 13.06 & 13.75 \\
     LMB & 45.47 & 46.26 & 46.24 & \textbf{47.60} & 46.13 & \underline{47.53} & 46.56 & 45.04 \\
     PIQA & 73.12 & 74.10 & 73.78 & 72.91 & \underline{74.37} & 72.91 & \textbf{74.37} & 74.10 \\
     Hella & 56.29 & 57.55 & \textbf{58.58} & 56.99 & 57.00 & \underline{58.47} & 56.85 & 57.17 \\
     Wino & 55.88 & 58.01 & \underline{58.48} & 57.22 & 57.85 & \textbf{60.14} & 57.38 & 57.62 \\
     ARC-e & 73.40 & 72.14 & 72.98 & 73.06 & 72.05 & \underline{74.28} & 73.11 & \textbf{74.28} \\
     ARC-c & \textbf{40.61} & 36.95 & 39.33 & \underline{40.36} & 39.76 & 40.02 & 39.33 & 39.33 \\
     SIQA & 40.74 & 41.71 & 41.81 & \underline{42.37} & 41.81 & \textbf{42.78} & 42.07 & 42.07 \\
     BoolQ & 60.58 & 53.98 & 60.52 & 62.45 & \underline{62.51} & 62.11 & \textbf{63.03} & 56.27 \\
      \midrule
      \rowcolor{gray!10}\multicolumn{9}{c}{\emph{ Generalization }} \\
     RACE & 34.45 & 33.78 & 32.15 & 35.22 & 34.55 & \textbf{35.60} & \underline{35.22} & 35.02 \\
     BBQ & 29.53 & 29.75 & 29.43 & 29.95 & 30.55 & \textbf{31.46} & \underline{30.88} & 30.27 \\
     MetaBench & 26.97 & 28.67 & 27.70 & 25.64 & \textbf{29.55} & 26.79 & \underline{29.38} & 28.98 \\
     QA4MRE & \textbf{40.00} & 35.00 & 39.17 & 39.17 & \underline{39.17} & 38.33 & 38.33 & 38.33 \\
     SCIQ & 89.80 & \underline{90.30} & 90.30 & 89.60 & 89.50 & 89.20 & \textbf{90.40} & 89.20 \\
     SWAG & 47.69 & 48.17 & \textbf{48.88} & 48.22 & 47.80 & \underline{48.57} & 48.18 & 47.80 \\
      \midrule
    \rowcolor{gray!10}\multicolumn{9}{c}{\emph{ Averages }} \\
     Dev. Avg & 55.76 & 55.09 & 56.47 & \underline{56.62} & 56.44 & \textbf{57.28} & 56.59 & 55.74 \\
     Gen. Avg & 44.74 & 44.28 & 44.61 & 44.63 & \underline{45.19} & 44.99 & \textbf{45.40} & 44.93 \\
      Overall Avg & 51.04 & 50.46 & 51.38 & 51.48 & 51.61 & \textbf{52.01} & \underline{51.79} & 51.11\\
      \bottomrule
    \end{tabular*}
  \caption{Top block: 10 development benchmarks, used in our exploration stage; bottom block: 6 generalization benchmarks for out-of-distribution testing. Dev., Gen., and Overall averages cover the 8 development, 6 generalization, and all 14 accuracy metrics, respectively; perplexity is reported separately. \textbf{Bold} indicates the best results, and \underline{underlining} indicates the second-best results. Model abbreviations are as follows: PG = PathGateFusionNet, C = ContentSharpRouter, FG = FusionGatedFIRNet, H = HierGateNet, and AM = AdaMultiPathGateNet.}
    \label{tab:id_ood_1p3b}
  \end{table*}

\FloatBarrier
\section{Analysis}

Throughout this section we track statistics in windows of 50 candidates (matching the pool update cadence) and refer to the top-105 architectures collectively as the \textit{model gallery}.

\subsection{Evolution dynamics and fitness design}
\label{sec:continuous-discovery}

\begin{figure*}[!ht]
  \centering
  \includegraphics[width=\linewidth]{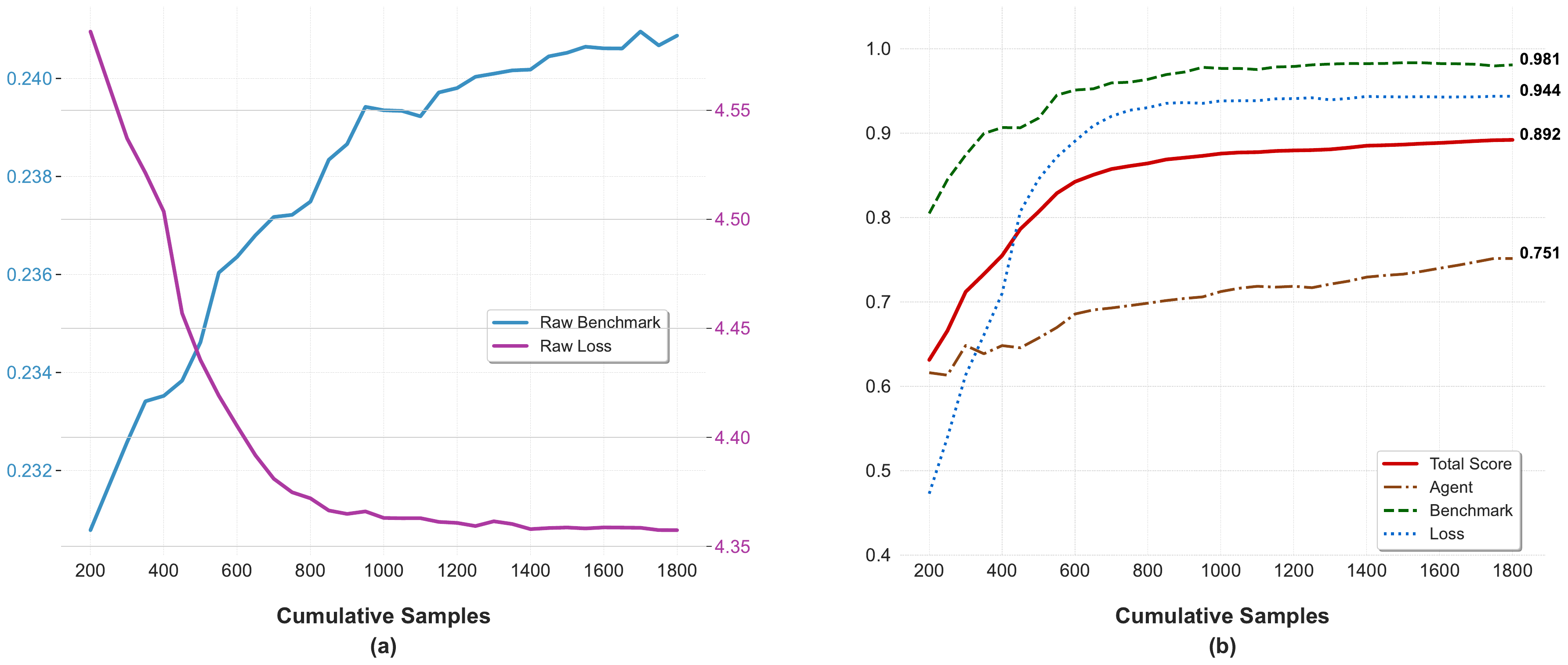}
  \caption{(a) Average raw benchmark score and loss of top-50 candidates vs.\ cumulative samples. (b) Composite fitness and its components vs.\ cumulative samples. The fitness plateau is a designed effect of the sigmoid transformation, not a performance ceiling.}
  \label{fig:annly_top}
  \vspace{-6pt}
\end{figure*}

Figure~\ref{fig:annly_top} reveals two complementary patterns.

\FloatBarrier

Figure~\ref{fig:annly_top}b shows that the composite fitness score of the top-50 candidate pool rises sharply over the first approximately 400 samples as the system quickly identifies architectures that outperform the DeltaNet baseline on both loss and benchmarks. Around 900 samples, however, the growth of the fitness score begins to slow. Comparing Figure~\ref{fig:annly_top}a and Figure~\ref{fig:annly_top}b reveals that this slowdown does not mean the search stops improving: the system continues to discover architectures with lower loss and higher benchmark scores, but the sigmoid transformation maps these raw gains into progressively smaller changes in the rule-based objective part of the fitness. We intentionally adopt this design so that no single metric can dominate the optimization signal indefinitely. As a result, the relative influence of the LLM judge becomes larger in the later stage of the search, encouraging the system to explore designs that are more principled, refined, and diverse rather than over-exploit a narrow performance regime. A direct piece of evidence is that the subjective score assigned by the LLM judge continues to improve steadily throughout the run.

Another key question is whether the design of the fitness function truly determines the system's evolutionary trajectory. Due to the sigmoid design, score differences among nodes become smaller after 900 samples, making the fitness landscape relatively flatter in the later stage of the search. We therefore examined the records of the 105 architectures in the model gallery and found that, among the architectures discovered after 900 samples, approximately 78\% have parent nodes from the first 900 samples. This shows that the fitness function successfully drives the system to effectively reuse previously discovered high-quality architectures in the later stage, forming a breadth-wise search over strong existing results rather than merely performing local optimization by repeatedly iterating on the same result.

To test whether these dynamics are a single-run artifact, we conducted two independent full-pipeline reruns with DeepSeek-V4-Flash under the 120-result configuration described in Appendix~\ref{sec:additional-results}. The runs reached final best scores of 2.5161 and 2.4311, while their area under the best-score curve (AUC) values differed by only 0.45\% (281.3209 vs.\ 280.0682), supporting the reproducibility of the evolution trend (Table~\ref{tab:reruns}).

\subsection{Knowledge sources and evolutionary shift}

\begin{figure*}[!t]
  \centering
  \begin{minipage}{0.50\textwidth}
    \centering
    \vspace{-1mm}
    \small
    \setlength{\tabcolsep}{1.2pt}
      \begin{tabular}{@{}l|ccc@{}}
      \toprule
      \multicolumn{1}{l}{\textbf{Category}} & \textbf{Experience} & \multicolumn{1}{c}{\textbf{Cognition}} & \multicolumn{1}{l}{\textbf{Originality}}\\
      \midrule
      Model Gallery    & 44.8\%  & 48.6\% & 6.6\%\\
      Others     & 37.7\%  & 51.9\% & 10.4\%\\
      All     & 38.2\%  & 51.7\% & 10.1\%\\
      \bottomrule
      \end{tabular}%
    \captionof{table}{Comparison of the influence of pipeline components on SOTA versus other model designs. The data reveal a greater dependence on empirical analysis in the development of SOTA architectures.}
    \label{tab:motivation_source}%
  \end{minipage}%
  \begin{minipage}{0.50\textwidth}
    \centering
    \vspace{0mm}
    \includegraphics[width=1.\textwidth]{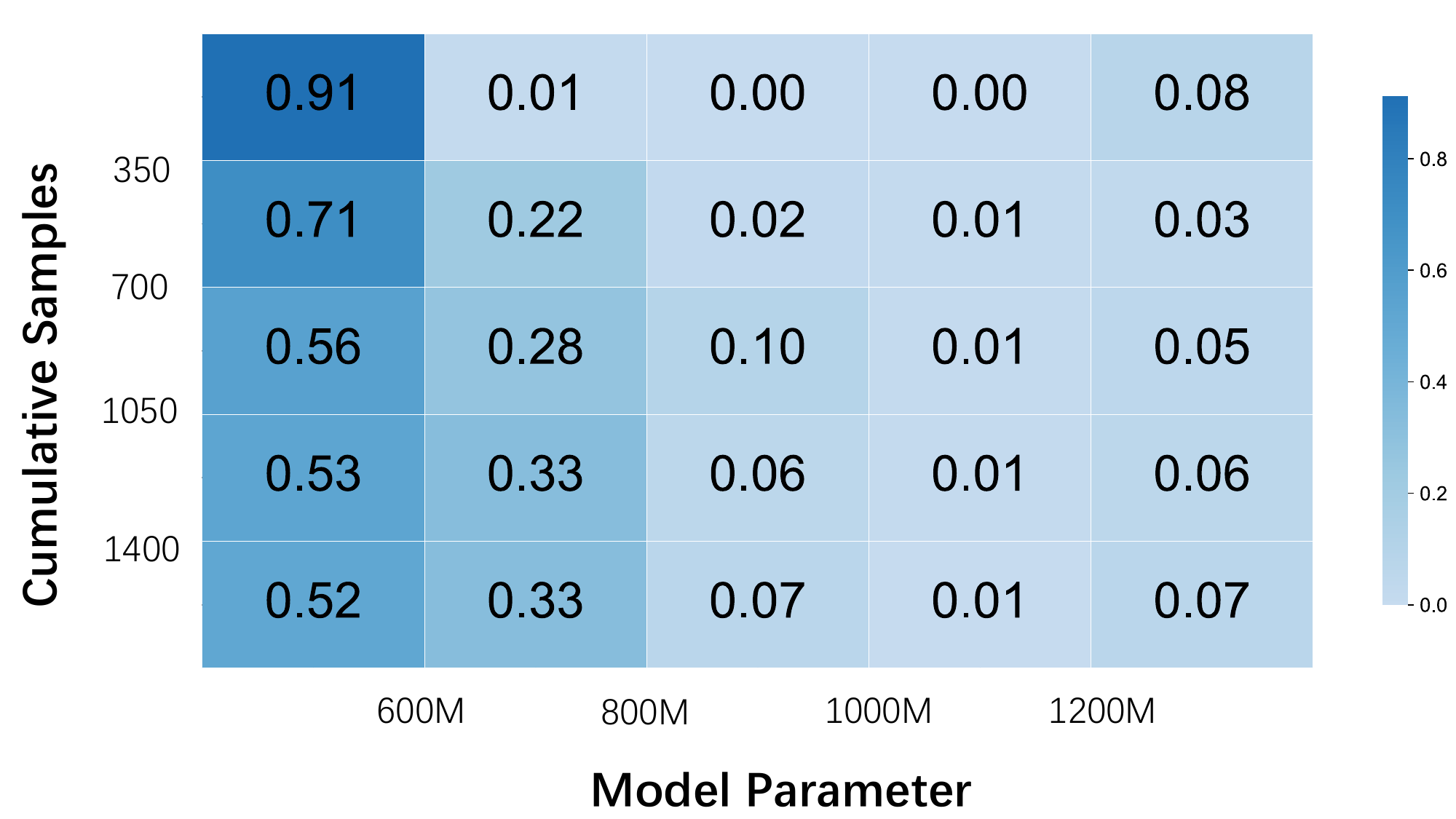}
    \vspace{-6mm}
    \captionof{figure}{Parameter-count distribution of candidate architectures.
    }
    \label{fig:exp2-walle-on-different-model-size}%
  \end{minipage}
\vspace{-3.5mm}
\end{figure*}

Each proposed architecture can draw on three sources of inspiration: \emph{cognition} (knowledge retrieved from the human literature base), \emph{experience} (insights synthesized from past experimental analyses), or \emph{originality} (novel ideas generated by the model itself). We prompted an LLM to classify the origin of each architectural component for both the full set of 1,773 architectures and the 105 architectures in the model gallery (Table~\ref{tab:motivation_source}).

Cognition is the dominant source throughout, accounting for 51.7\% of all design ideas and remaining at 48.6\% within the model gallery---confirming that human prior knowledge is indispensable for producing technically grounded proposals. At the same time, a meaningful shift emerges among top performers: experience rises from 38.2\% to 44.8\%, echoing the pattern in human scientific progress where breakthroughs come not from prior literature alone but from synthesizing insights accumulated through one's own experimentation. Finally, originality remains low throughout: 10.1\% in the full set and 6.6\% in the model gallery. This suggests that the system usually grounds its proposals in the knowledge base rather than relying on ideas generated from scratch. In this highly specialized domain, that is exactly the behavior one would want.

To test whether cognition preparation can also be automated, we built a cognition source using GPT-5.5 with Deep Research and ran the same evolve pipeline. Across the first 120 results, the AI-curated run reached a final best raw score of 2.5500 and an AUC of 285.6003, while the matched human-curated run reached a final best raw score of 2.5161 and an AUC of 281.3209 (Table~\ref{tab:ai-cognition}). These comparable run-level outcomes provide evidence that an AI-curated cognition source can support \modelname without extensive manual curation.

\subsection{Architectural design patterns}

\FloatBarrier
\begin{figure}[!ht]
  \centering
  \includegraphics[width=\linewidth]{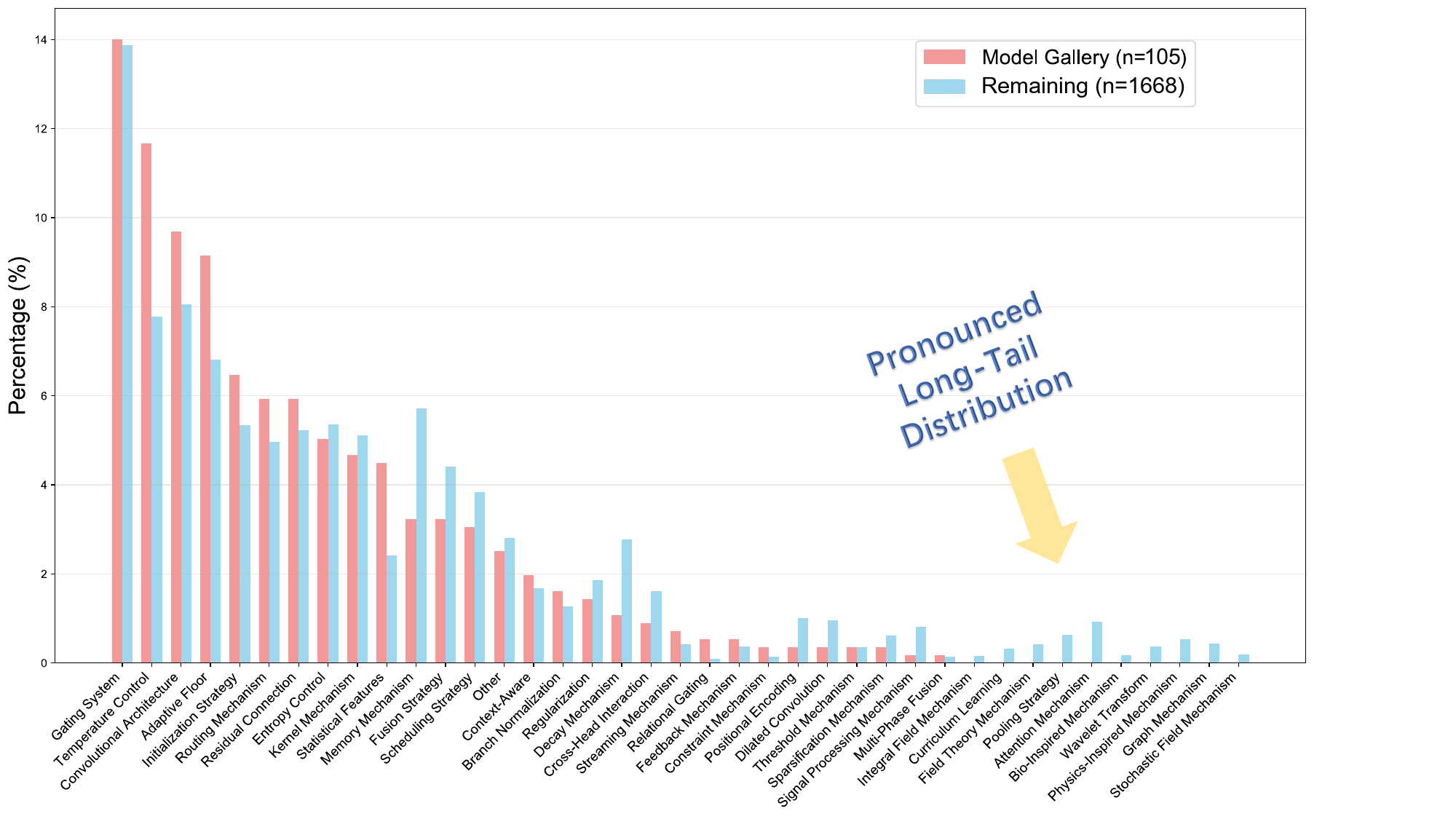}
  \caption{Component-mention distributions: model gallery (top 105) vs.\ remaining architectures.}
  \label{fig:comp_analy}
  \vspace{-6pt}
\end{figure}

We examined the parameter count distribution of candidate architectures under a common reference configuration across iterations (Figure~\ref{fig:exp2-walle-on-different-model-size}). Most candidates remained within the lower parameter ranges throughout the search, suggesting that performance gains were not driven solely by parameter growth. To understand what that innovation looks like in practice, we parsed all 1,773 design motivations with an auxiliary LLM, extracting over 5,000 component mentions and grouping them into high-level categories (Figure~\ref{fig:comp_analy}); percentages denote each category's share of component mentions within the corresponding group. The model gallery shows a notably less pronounced long-tail distribution than the remaining architectures: top models converge on a core set of proven components---gating mechanisms and convolutions---and rarely adopt exotic alternatives such as physics-inspired mechanisms. This pattern mirrors human scientific practice, where state-of-the-art results are more reliably achieved through targeted innovation on a foundation of validated techniques than through the pursuit of novelty for its own sake.

\subsection{Cross-family transfer}

To more comprehensively assess the quality of the architectures discovered by \modelname, we investigate whether the design principles learned on DeltaNet generalize beyond the family in which they were discovered---a stronger test of their value than within-family performance alone. The system summarized regularities from the top 105 DeltaNet variants into transferable design rules and prompts specifying architectural rewrites and implementation constraints. The system applies these rules to Mamba2 (SSM), GLA, and HGRN2, synthesizing candidate variants for each family. All candidates pass the same novelty, sanity, and efficiency checks (Sections~\ref{sec:evolution} and~\ref{sec:validation}) and are trained under identical configurations.

\begin{table}[!ht]
  \centering
  \small
  \setlength{\tabcolsep}{6pt}
  \begin{tabular}{lccc}
    \toprule
    \textbf{Family} & \textbf{Baseline avg.} & \textbf{Variant avg.} & \textbf{Gain} \\
    \midrule
    Mamba2 & 47.84 & 48.19 & +0.35 \\
    HGRN2  & 47.67 & 47.81 & +0.14 \\
    GLA    & 45.55 & 47.60 & +2.05 \\
    \bottomrule
  \end{tabular}
  \caption{Summary of cross-family transfer at 340M parameters. Each row reports the baseline and the selected variant; full benchmark results are reported in Appendix Table~\ref{tab:cross_family_full}.}
  \label{tab:cross_family_mid}
\end{table}

\paragraph{Transferability of distilled rules.}
Although the Delta-rule algorithm itself is difficult to transfer directly across model families, rules centered on routing, gating, and stability transfer effectively to Mamba2, GLA, and HGRN2, yielding improvements over the corresponding baseline for each target family (Table~\ref{tab:cross_family_mid}). This suggests that the discovered principles capture generalizable architectural behaviors rather than DeltaNet-specific quirks.

\paragraph{Per-family outcomes.}
GLA shows the most robust transfer. Mamba2 and HGRN2 also achieve performance gains over their baselines, but the improvements are relatively smaller. We attribute this to two factors: (1) \textbf{Structural alignment}---DeltaNet and GLA share linearized attention-style routing and gating, so DeltaNet-derived rules map more naturally onto GLA's computation graph; (2) \textbf{Baseline saturation and interaction effects}---Mamba2 and HGRN2 are already heavily optimized and incorporate multiple stabilization mechanisms; external rule insertions may partially overlap with or conflict with their intrinsic mechanisms, producing diminishing returns. Future work will develop family-specific adapters to reduce interference.

\subsection{Efficiency at large scale}

We measure end-to-end efficiency under the same local implementation and hardware setting. Training throughput includes the forward pass, language-modeling loss, backward pass, and optimizer step; prefill measures full-sequence inference with cache enabled. As shown in Table~\ref{tab:efficiency-1p3b}, the discovered architectures maintain training throughput in the same practical range as the baselines, while several achieve prefill throughput comparable to Mamba2. This efficiency remains competitive despite the additional computation introduced by routing, gating, and extra projections in some finalists.

\begin{table}[!ht]
  \centering
  \small
  \begin{tabular}{lrrr}
    \toprule
    \textbf{Model} & \textbf{Train tok/s} $\uparrow$ & \textbf{Prefill tok/s} $\uparrow$ & \textbf{Peak mem.} $\downarrow$ \\
    \midrule
    DeltaNet & 5,063 & 23,162 & 11.44G \\
    Gated DeltaNet & 4,959 & 22,413 & 11.54G \\
    Mamba2 & 4,476 & 12,725 & 11.29G \\
    \midrule
    \textsc{ASI-Arch}-1 & 4,037 & 13,589 & 18.32G \\
    \textsc{ASI-Arch}-2 & 4,168 & 12,237 & 16.28G \\
    \textsc{ASI-Arch}-3 & 3,426 & 11,932 & 18.39G \\
    \textsc{ASI-Arch}-4 & 4,027 & 12,950 & 22.75G \\
    \textsc{ASI-Arch}-5 & 3,923 & 11,936 & 16.57G \\
    \bottomrule
  \end{tabular}
  \caption{End-to-end efficiency at 1.3B scale under the same implementation and hardware setting.}
  \label{tab:efficiency-1p3b}
\end{table}
\FloatBarrier

\subsection{Framework comparison}

We compare \modelname with OpenEvolve~\citep{openevolve2025} using the same DeepSeek-V4-Flash proposal model and a 120-attempt budget; candidate architectures are trained under the 20M/1B-token configuration. Because the systems use different native scores, Table~\ref{tab:openevolve} reports metric-wise best values across successful candidates: Loss is the lowest final-step training loss, while Benchmark is the highest mean accuracy over LAMBADA, HellaSwag, PIQA, WinoGrande, ARC-Easy, ARC-Challenge, Social-IQA, and BoolQ. The two values need not come from the same candidate. The rate is the fraction of raw attempts that produce a database result satisfying the architecture requirements. Parenthesized percentages are improvements relative to the DeltaNet baseline (loss 4.3861; benchmark 0.3205).

\begin{table}[!ht]
  \centering
  \small
  \begin{tabular}{lccc}
    \toprule
    \textbf{Strategy} & \textbf{Rate} & \textbf{Loss} $\downarrow$ & \textbf{Benchmark} $\uparrow$ \\
    \midrule
    \modelname & 37.4\% & 4.1043 (6.4\%) & 0.3500 (9.2\%) \\
    OpenEvolve & 8.3\% & 4.3454 (0.9\%) & 0.3405 (6.2\%) \\
    \bottomrule
  \end{tabular}
  \caption{Strategy-level comparison under the same proposal model and attempt budget.}
  \label{tab:openevolve}
\end{table}
\FloatBarrier

Under this matched setting, \modelname achieves approximately $4.5\times$ the successful-design rate of OpenEvolve together with stronger loss and benchmark improvements, showing that its task-specific pipeline provides a substantial advantage over a generic evolution strategy for architecture discovery.

\section{Conclusion}

We presented \modelname, an autonomous system for AI-driven neural architecture research that operates through a closed-loop \emph{research-experiment-analyze-update} process without human intervention. In linear attention, the system continuously discovered SOTA architectures throughout 1,773 experiments, yielding 105 novel designs with no saturation in sight. Distilled design rules further transferred across families from DeltaNet to Mamba2, GLA, and HGRN2. We analyzed the design patterns and fitness dynamics that enabled these results and open-sourced the framework and all discovered architectures. Future work will explore larger architectural hypothesis spaces, continue iterating on the strong designs discovered here, and pursue concrete algorithmic efficiency optimizations to better understand their practical efficiency.

\appendix
\section{LLM disclosure}
\label{sec:llm-disclosure}

We disclose major uses of large language models in this work. Experiments used OpenAI's \textbf{o3}, \textbf{GPT-4.1}, and \textbf{GPT-5.5 with Deep Research}, as well as \textbf{DeepSeek-V4-Flash}, across the \modelname pipeline and additional analyses. We also used LLMs to assist with edits and refactoring in the experiment codebase. The manuscript was written by the authors; LLM assistance in paper preparation was limited to non-substantive LaTeX and table formatting, and no LLMs were used to draft scientific content.

\section{Experimental setup}
\label{sec:mid-eval}

\subsection{Experimental configuration}

\paragraph{Baseline and fitness calibration.}
We use DeltaNet~\citep{yang2024parallelizing} as our primary baseline. For quantitative fitness scoring, we map performance differences within 10\% of baseline through a sigmoid function; architectures whose training loss falls more than 10\% below baseline are immediately discarded as likely information leakage. For qualitative scoring, DeltaNet is set at 5 points and Gated DeltaNet~\citep{yang2024gated} at 10 points; the LLM evaluator scores each candidate on this 1--10 scale, rescaled to $[0,1]$ via $(s-1)/9$.

\paragraph{Parallel search and pool management.}
Multiple search processes run in parallel, coordinated through a shared cloud database. The first 200 explorations proceed without pool updates (cold start), after which the candidate pool is updated every 50 new entries (batched updates), ensuring all agents reference a stable set during each exploration batch.

\paragraph{Two-stage evaluation.}
\begin{itemize}[leftmargin=1.5em, labelindent=0.5em, itemsep=3pt, parsep=0pt, topsep=3pt]
\item \textbf{Exploration stage}: 20M parameters (8 layers, hidden dim 256, 8 heads), 2,000 steps on 1B FineWeb-Edu tokens, 500 samples per benchmark.
\item \textbf{Verification stage}: Promising candidates scaled to 340M (24 layers, hidden dim 1024), trained on 1B tokens with a strict causality check. Top models further scaled to 1.3B and trained for 30,000 steps on 15B tokens, evaluated on 16 benchmarks.
\end{itemize}

\subsection{Pipeline configuration}

\paragraph{Framework overview.}
Our experimental framework implements an automated AI self-iterative system for exploring novel neural network architectures through three core phases: Evolve, Training, and Analysis. The system operates cyclically, extracting nodes from a MongoDB database, generating new motivations and implementations, conducting training and evaluation, and performing comprehensive analysis with knowledge integration.

\paragraph{Multi-model integration.}
We employ a hybrid multi-model approach to optimize both quality and efficiency. In the Evolve phase, we combine o3 and GPT-4.1 models for the planner component to balance motivation quality and generation speed while enhancing architectural diversity. The checker component utilizes o3 to ensure code validity and prevent resource waste, while motivation deduplication employs GPT-4.1 for rapid processing. During the Training phase, GPT-4.1 handles training initiation, testing, and debugging operations, focusing on detail-level modifications without structural changes for rapid iteration capabilities. The Analysis phase utilizes o3 to conduct comprehensive experimental analysis, providing high-quality insights to enhance subsequent exploration efficiency.

\paragraph{Data management and retrieval.}
For data management and retrieval, MongoDB serves as our primary storage solution, supporting name-based and sequential storage along with deletion functionality for experimental nodes. FAISS enables efficient similarity matching during motivation deduplication, identifying similar concepts in the database before agent-based verification to improve exploration efficiency. We curate cognitive insights from relevant literature and index them in OpenSearch for retrieval-augmented generation. For each experimental result, we retrieve the three most similar cognitive entries and integrate them into the experimental node to support future exploration.

\subsection{Model and training details}

\paragraph{Progressive evaluation strategy.}
To balance exploration efficiency with computational constraints, we implement a three-tiered progressive evaluation approach with rapid architecture exploration using 20M parameter models, followed by validation phases at larger scales.

\paragraph{Model architecture.}
The base 20M configuration employs 8 attention heads across 8 hidden layers with a hidden dimension of 256, maintaining computational tractability while preserving architectural expressiveness. The model uses SiLU activation for query-key transformations with L2 normalization and incorporates short convolutions with a kernel size of 4. For comparison, 340M parameter models employ 1024 hidden size, 8 attention heads, and 24 hidden layers with tied word embeddings disabled.

\paragraph{Training protocol.}
We utilize the FLAME framework~\citep{yang2025flame} with AdamW optimization, employing a peak learning rate of $3 \times 10^{-4}$, epsilon value of $1 \times 10^{-8}$, and a warmup-stabilize-decay (WSD) learning rate schedule. Training proceeds for 2,000 steps for 20M models, using mixed precision training with bfloat16 parameters and float32 gradient reduction. All models maintain a consistent batch size of 256 and employ GPT-2 tokenizer throughout training and evaluation phases.

\paragraph{Data configuration.}
Training utilizes FineWeb-Edu sample-10BT and sample-100BT datasets~\citep{penedo2024fineweb} with a context length of 2048 tokens. For comprehensive evaluation of discovered architectures, we scale to 340M parameter models that provide more reliable performance assessment. The 340M models are trained using 1 billion tokens. The 1.3B models are trained using 15 billion tokens from sample-100BT, with a learning-rate warm-up phase of approximately 0.5 billion tokens.

\paragraph{Evaluation protocol.}
Model evaluation is performed using the LM-Evaluation-Harness framework~\citep{eval-harness}, a standardized open-source tool developed by EleutherAI that provides unified benchmarking protocols for language models. The evaluation suite assesses a diverse range of capabilities, including perplexity (Wiki PPL, LAMBADA PPL)~\citep{merity2016pointer, paperno2016lambada}, reading comprehension (LAMBADA acc, RACE acc)~\citep{paperno2016lambada, lai-etal-2017-race}, commonsense reasoning (PIQA acc, HellaSwag acc, SWAG acc, Social-IQA acc, WinoGrande acc)~\citep{Bisk2020, zellers2019hellaswag, zellers2018swagaf, sap2019social, sakaguchi2019winogrande}, knowledge-intensive tasks (ARC-Easy acc, ARC-Challenge acc, SciQ acc, QA4MRE-2011 acc)~\citep{Clark2018ThinkYH, Welbl2017CrowdsourcingMC, Peas2013QA4MRE2O}, and reasoning and general capabilities (BoolQ acc, BBQ acc, MetaBench acc)~\citep{clark2019boolq, parrish-etal-2022-bbq, metabench}. For rapid architectural exploration, the 20M parameter models are evaluated on a limited sample size of 500 per dataset, while validation phases use full datasets for all other models. All evaluations are conducted with consistent hyperparameters to ensure fair comparison across architectural variants: a `dtype' of bfloat16, a `max\_length' of 4096, and model-specific batch sizes of 64 for the 20M models, 4 for the 340M models, and 8 for the 1.3B models. Final model ranking is a comprehensive assessment that incorporates LLM subjective evaluation, training loss metrics, and benchmark performance.

\subsection{Additional reproducibility and cognition-source results}
\label{sec:additional-results}

\paragraph{Reproducibility across independent runs.}
We reran the complete evolve pipeline twice with DeepSeek-V4-Flash, with each run producing 120 results. For these reruns, we removed the 200-exploration frozen-pool cold start and refreshed the candidate pool every 20 results. The similar best-score dynamics in Table~\ref{tab:reruns} indicate that the evolution trend is not a single-run artifact. In Tables~\ref{tab:reruns} and~\ref{tab:ai-cognition}, the reported raw total score is the unnormalized sum of the three fitness components before the factor of $1/3$ in Equation~(1); AUC and curve-deviation statistics are computed from this raw-total-score curve. AUC denotes the area under the curve over the first 120 results. For both reruns, the standard deviation from the average curve is 0.0798, corresponding to 3.41\% of the mean raw total score along the curve.

\begin{table}[!ht]
  \centering
  \small
  \begin{tabular}{lccc}
    \toprule
    \textbf{Metric} & \textbf{Run 1} & \textbf{Run 2} & \textbf{Difference} \\
    \midrule
    Final best score & 2.5161 & 2.4311 & 0.0850 (3.44\%) \\
    AUC & 281.3209 & 280.0682 & 1.2527 (0.45\%) \\
    Std. dev. from average curve & 0.0798 & 0.0798 & -- \\
    \bottomrule
  \end{tabular}
  \caption{Two independent full-pipeline reruns.}
  \label{tab:reruns}
\end{table}

\paragraph{AI-curated cognition source.}
To test whether \modelname requires a manually curated cognition base, we constructed a cognition source from the output of GPT-5.5 with Deep Research. We asked the model to identify papers related to linear attention and neural architecture design and return their metadata, summaries, and relevance. We converted the report into the same cognition schema used by \modelname. Table~\ref{tab:ai-cognition} compares this run with the first reproducibility rerun in Table~\ref{tab:reruns}, which uses the original human-curated cognition base. Both runs used DeepSeek-V4-Flash, omitted the 200-exploration frozen-pool cold start, refreshed the candidate pool every 20 results, and are reported over their first 120 results.

\begin{table}[!ht]
  \centering
  \small
  \begin{tabular}{lcc}
    \toprule
    \textbf{Metric} & \textbf{Human Curated} & \textbf{AI Curated} \\
    \midrule
    Final best score & 2.5161 & 2.5500 \\
    AUC & 281.3209 & 285.6003 \\
    Best score at 20 results & 2.3518 & 2.3526 \\
    Best score at 40 results & 2.3518 & 2.3526 \\
    Best score at 60 results & 2.3518 & 2.4510 \\
    Best score at 80 results & 2.3518 & 2.5500 \\
    Best score at 100 results & 2.3518 & 2.5500 \\
    Best score at 120 results & 2.5161 & 2.5500 \\
    \bottomrule
  \end{tabular}
  \caption{Comparison of human-curated and AI-curated cognition sources over the first 120 results.}
  \label{tab:ai-cognition}
\end{table}

\subsection{Supplementary results}

Table~\ref{tab:cross_family_full} reports the complete benchmark results for cross-family transfer, while Table~\ref{tab:model_comparison_107} displays the 20M/340M results for 105 discovered architectures. Some discovered architectures share the same generated name and exhibit similar designs. We group these entries under the same name and report their distinct 340M results on separate rows.

\begin{table*}[!ht]
    \centering
    \small
    \renewcommand{\arraystretch}{0.95}
    \setlength{\tabcolsep}{1.5pt}
    \begin{tabular}{@{}lccccccccccc@{}}
      \toprule
      \textbf{Model} & \textbf{Wiki} $\downarrow$ & \textbf{LMB} $\downarrow$ & \textbf{LMB} $\uparrow$ & \textbf{PIQA} $\uparrow$ & \textbf{Hella} $\uparrow$ & \textbf{Wino} $\uparrow$ & \textbf{ARC-e} $\uparrow$ & \textbf{ARC-c} $\uparrow$ & \textbf{SIQA} $\uparrow$ & \textbf{BoolQ} $\uparrow$ & \textbf{Avg.} \\
      \midrule
      Mamba2 & 27.08 & 40.09 & 31.32 & \textbf{67.90} & 42.25 & \textbf{51.46} & 62.04 & 29.27 & \textbf{39.25} & 59.24 & 47.84 \\
      \makecell[l]{Mamba2\\variant} & \textbf{26.79} & \textbf{38.46} & \textbf{32.49} & 67.52 & \textbf{42.40} & 51.30 & \textbf{63.22} & \textbf{30.29} & 38.49 & \textbf{59.79} & \textbf{48.19} \\
      \midrule
      HGRN2 & 28.09 & \textbf{36.02} & 32.39 & \textbf{68.82} & \textbf{41.42} & \textbf{51.30} & 61.03 & 29.01 & 37.92 & 59.48 & 47.67 \\
      \makecell[l]{HGRN2\\variant} & \textbf{27.56} & 38.31 & \textbf{32.45} & 68.28 & 41.33 & 51.07 & 61.03 & \textbf{29.14} & \textbf{38.95} & \textbf{60.21} & \textbf{47.81} \\
      \midrule
      GLA & 30.83 & 53.44 & 28.55 & 65.61 & 38.04 & \textbf{51.14} & 57.87 & 26.54 & 37.51 & 59.14 & 45.55 \\
      \makecell[l]{GLA\\variant} & \textbf{27.48} & \textbf{36.08} & \textbf{32.84} & \textbf{67.52} & \textbf{41.51} & 50.67 & \textbf{62.21} & \textbf{28.16} & \textbf{38.59} & \textbf{59.33} & \textbf{47.60} \\
      \bottomrule
    \end{tabular}
    \caption{Complete cross-family transfer results at 340M parameters. For each family, we report the baseline and the selected variant. Metrics follow the main setup: Wiki ppl, LAMBADA ppl/acc, PIQA, HellaSwag (normalized), WinoGrande, ARC-Easy, ARC-Challenge (normalized), SIQA, BoolQ, and the average across accuracy metrics. \textbf{Bold} indicates the better result within each family pair.}
    \label{tab:cross_family_full}
  \end{table*}

\clearpage

\setlength{\LTleft}{\fill}
\setlength{\LTright}{\fill}
\setlength{\tabcolsep}{3.5pt}
\footnotesize
\begin{longtable}{p{0.40\linewidth}|cc|cc}
\toprule
\multirow{2}{*}{\textbf{Model Name}} & \multicolumn{2}{c|}{\textit{20M params / 1B tokens}} & \multicolumn{2}{c}{\textit{340M params / 1B tokens}} \\
\cmidrule(lr){2-3} \cmidrule(lr){4-5}
& \textbf{Train Loss $\downarrow$} & \textbf{Test Score $\uparrow$} & \textbf{Train Loss $\downarrow$} & \textbf{Test Score $\uparrow$} \\
\midrule
\endfirsthead

\multicolumn{5}{c}%
{{\tablename\ \thetable{} -- continued from previous page}} \\
\toprule
\multirow{2}{*}{\textbf{Model Name}} & \multicolumn{2}{c|}{\textit{20M params / 1B tokens}} & \multicolumn{2}{c}{\textit{340M params / 1B tokens}} \\
\cmidrule(lr){2-3} \cmidrule(lr){4-5}
& \textbf{Train Loss $\downarrow$} & \textbf{Test Score $\uparrow$} & \textbf{Train Loss $\downarrow$} & \textbf{Test Score $\uparrow$} \\
\midrule
\endhead

\midrule \multicolumn{5}{r}{{Continued on next page}} \\
\endfoot

\bottomrule
\caption{Model performance comparison. Train Loss represents the loss at the final training step. Test Score is the average performance across 7 tasks: ARC-Challenge, ARC-Easy, BoolQ, HellaSwag, PIQA, Social IQA, and WinoGrande. Green subscripts indicate improvements over the Gated DeltaNet baseline.}\label{tab:model_comparison_107} \\
\endlastfoot

\rowcolor{gray!10} DeltaNet (Baseline) & 4.5749 & 36.23 & 3.5055 & 41.16 \\
\rowcolor{gray!10} Gated DeltaNet (Baseline) & 4.5678 & 36.60 & 3.4768 & 42.10 \\
AdaptiveContextFusionNet & 4.4973 $_{\textcolor{ForestGreen}{-0.0705}}$ & 37.03 $_{\textcolor{ForestGreen}{+0.43}}$ & 3.4624 $_{\textcolor{ForestGreen}{-0.0144}}$ & 42.74 $_{\textcolor{ForestGreen}{+0.64}}$ \\
AdaptiveEntropyGateNet & 4.4423 $_{\textcolor{ForestGreen}{-0.1255}}$ & 36.91 $_{\textcolor{ForestGreen}{+0.31}}$ & 3.4558 $_{\textcolor{ForestGreen}{-0.0210}}$ & 42.37 $_{\textcolor{ForestGreen}{+0.27}}$ \\
AdaptiveEntropyRouter & 4.3547 $_{\textcolor{ForestGreen}{-0.2131}}$ & 39.26 $_{\textcolor{ForestGreen}{+2.66}}$ & 3.4066 $_{\textcolor{ForestGreen}{-0.0702}}$ & 44.31 $_{\textcolor{ForestGreen}{+2.21}}$ \\
AdaptiveEntropyRouterNet & 4.3326 $_{\textcolor{ForestGreen}{-0.2352}}$ & 36.94 $_{\textcolor{ForestGreen}{+0.34}}$ & 3.4298 $_{\textcolor{ForestGreen}{-0.0470}}$ & 43.25 $_{\textcolor{ForestGreen}{+1.15}}$ \\
AdaptiveFloorGate & 4.4695 $_{\textcolor{ForestGreen}{-0.0983}}$ & 37.00 $_{\textcolor{ForestGreen}{+0.40}}$ & 3.4418 $_{\textcolor{ForestGreen}{-0.0350}}$ & 43.57 $_{\textcolor{ForestGreen}{+1.47}}$ \\
AdaptiveFloorNet-HAF & 4.4002 $_{\textcolor{ForestGreen}{-0.1676}}$ & 37.03 $_{\textcolor{ForestGreen}{+0.43}}$ & 3.4241 $_{\textcolor{ForestGreen}{-0.0527}}$ & 43.59 $_{\textcolor{ForestGreen}{+1.49}}$ \\
AdaptiveFractalGateNet & 4.5484 $_{\textcolor{ForestGreen}{-0.0194}}$ & 38.43 $_{\textcolor{ForestGreen}{+1.83}}$ & 3.4351 $_{\textcolor{ForestGreen}{-0.0417}}$ & 43.84 $_{\textcolor{ForestGreen}{+1.74}}$ \\
AdaptiveFusionNet & 4.3521 $_{\textcolor{ForestGreen}{-0.2157}}$ & 37.51 $_{\textcolor{ForestGreen}{+0.91}}$ & 3.4336 $_{\textcolor{ForestGreen}{-0.0432}}$ & 43.78 $_{\textcolor{ForestGreen}{+1.68}}$ \\
 & -- & -- & 3.4061 $_{\textcolor{ForestGreen}{-0.0707}}$ & 44.08 $_{\textcolor{ForestGreen}{+1.98}}$ \\
 & -- & -- & 3.4280 $_{\textcolor{ForestGreen}{-0.0488}}$ & 43.66 $_{\textcolor{ForestGreen}{+1.56}}$ \\
AdaptiveFusionNet-DSI & 4.3781 $_{\textcolor{ForestGreen}{-0.1897}}$ & 36.63 $_{\textcolor{ForestGreen}{+0.03}}$ & 3.4270 $_{\textcolor{ForestGreen}{-0.0498}}$ & 43.39 $_{\textcolor{ForestGreen}{+1.29}}$ \\
AdaptiveFusionRNet & 4.3940 $_{\textcolor{ForestGreen}{-0.1738}}$ & 37.03 $_{\textcolor{ForestGreen}{+0.43}}$ & 3.4086 $_{\textcolor{ForestGreen}{-0.0682}}$ & 43.73 $_{\textcolor{ForestGreen}{+1.63}}$ \\
AdaptiveGateNet & 4.4228 $_{\textcolor{ForestGreen}{-0.1450}}$ & 36.57 $_{-0.03}$ & 3.4377 $_{\textcolor{ForestGreen}{-0.0391}}$ & 43.64 $_{\textcolor{ForestGreen}{+1.54}}$ \\
AdaptiveGateNet-AFP & 4.4198 $_{\textcolor{ForestGreen}{-0.1480}}$ & 37.80 $_{\textcolor{ForestGreen}{+1.20}}$ & 3.4193 $_{\textcolor{ForestGreen}{-0.0575}}$ & 43.88 $_{\textcolor{ForestGreen}{+1.78}}$ \\
AdaptiveGateRouter\_X & 4.4126 $_{\textcolor{ForestGreen}{-0.1552}}$ & 38.69 $_{\textcolor{ForestGreen}{+2.09}}$ & 3.4114 $_{\textcolor{ForestGreen}{-0.0654}}$ & 42.68 $_{\textcolor{ForestGreen}{+0.58}}$ \\
AdaptiveGatedRouter-Hybrid & 4.3335 $_{\textcolor{ForestGreen}{-0.2343}}$ & 37.74 $_{\textcolor{ForestGreen}{+1.14}}$ & 3.4060 $_{\textcolor{ForestGreen}{-0.0708}}$ & 43.56 $_{\textcolor{ForestGreen}{+1.46}}$ \\
AdaptiveHierGateNet & 4.5096 $_{\textcolor{ForestGreen}{-0.0582}}$ & 36.71 $_{\textcolor{ForestGreen}{+0.11}}$ & 3.4433 $_{\textcolor{ForestGreen}{-0.0335}}$ & 43.44 $_{\textcolor{ForestGreen}{+1.34}}$ \\
AdaptiveHybridGateNet & 4.3867 $_{\textcolor{ForestGreen}{-0.1811}}$ & 37.11 $_{\textcolor{ForestGreen}{+0.51}}$ & 3.4239 $_{\textcolor{ForestGreen}{-0.0529}}$ & 43.01 $_{\textcolor{ForestGreen}{+0.91}}$ \\
AdaptiveMixGateNet & 4.3709 $_{\textcolor{ForestGreen}{-0.1969}}$ & 36.91 $_{\textcolor{ForestGreen}{+0.31}}$ & 3.4289 $_{\textcolor{ForestGreen}{-0.0479}}$ & 43.28 $_{\textcolor{ForestGreen}{+1.18}}$ \\
AdaptiveMixTransformer & 4.4855 $_{\textcolor{ForestGreen}{-0.0823}}$ & 36.49 $_{-0.11}$ & 3.4593 $_{\textcolor{ForestGreen}{-0.0175}}$ & 42.50 $_{\textcolor{ForestGreen}{+0.40}}$ \\
AdaptivePathRouter & 4.4324 $_{\textcolor{ForestGreen}{-0.1354}}$ & 38.00 $_{\textcolor{ForestGreen}{+1.40}}$ & 3.4332 $_{\textcolor{ForestGreen}{-0.0436}}$ & 42.42 $_{\textcolor{ForestGreen}{+0.32}}$ \\
AdaptiveSpanGateConv & 4.4431 $_{\textcolor{ForestGreen}{-0.1247}}$ & 36.74 $_{\textcolor{ForestGreen}{+0.14}}$ & 3.4247 $_{\textcolor{ForestGreen}{-0.0521}}$ & 43.90 $_{\textcolor{ForestGreen}{+1.80}}$ \\
AdaptiveTokenGate & 4.4954 $_{\textcolor{ForestGreen}{-0.0724}}$ & 36.77 $_{\textcolor{ForestGreen}{+0.17}}$ & 3.4400 $_{\textcolor{ForestGreen}{-0.0368}}$ & 44.08 $_{\textcolor{ForestGreen}{+1.98}}$ \\
AdaptiveTokenRouter & 4.3745 $_{\textcolor{ForestGreen}{-0.1933}}$ & 38.63 $_{\textcolor{ForestGreen}{+2.03}}$ & 3.4238 $_{\textcolor{ForestGreen}{-0.0530}}$ & 42.56 $_{\textcolor{ForestGreen}{+0.46}}$ \\
AnnealedPathFusionNet & 4.4564 $_{\textcolor{ForestGreen}{-0.1114}}$ & 36.83 $_{\textcolor{ForestGreen}{+0.23}}$ & 3.4472 $_{\textcolor{ForestGreen}{-0.0296}}$ & 43.73 $_{\textcolor{ForestGreen}{+1.63}}$ \\
BAMG\_MemoryGate & 4.4804 $_{\textcolor{ForestGreen}{-0.0874}}$ & 36.17 $_{-0.43}$ & 3.4587 $_{\textcolor{ForestGreen}{-0.0181}}$ & 43.42 $_{\textcolor{ForestGreen}{+1.32}}$ \\
BlockStateFusionNet & 4.3551 $_{\textcolor{ForestGreen}{-0.2127}}$ & 37.66 $_{\textcolor{ForestGreen}{+1.06}}$ & 3.4085 $_{\textcolor{ForestGreen}{-0.0683}}$ & 43.20 $_{\textcolor{ForestGreen}{+1.10}}$ \\
BoundedTempAnnealNet & 4.4331 $_{\textcolor{ForestGreen}{-0.1347}}$ & 39.60 $_{\textcolor{ForestGreen}{+3.00}}$ & 3.4098 $_{\textcolor{ForestGreen}{-0.0670}}$ & 43.19 $_{\textcolor{ForestGreen}{+1.09}}$ \\
ContentSharpRouter & 4.3127 $_{\textcolor{ForestGreen}{-0.2551}}$ & 37.00 $_{\textcolor{ForestGreen}{+0.40}}$ & 3.4229 $_{\textcolor{ForestGreen}{-0.0539}}$ & 43.42 $_{\textcolor{ForestGreen}{+1.32}}$ \\
ConvFusionWide31 & 4.4279 $_{\textcolor{ForestGreen}{-0.1399}}$ & 38.60 $_{\textcolor{ForestGreen}{+2.00}}$ & 3.4273 $_{\textcolor{ForestGreen}{-0.0495}}$ & 43.47 $_{\textcolor{ForestGreen}{+1.37}}$ \\
ConvexBlendFloorNet & 4.4021 $_{\textcolor{ForestGreen}{-0.1657}}$ & 37.80 $_{\textcolor{ForestGreen}{+1.20}}$ & 3.4265 $_{\textcolor{ForestGreen}{-0.0503}}$ & 43.54 $_{\textcolor{ForestGreen}{+1.44}}$ \\
DepthwiseConvPointMixer & 4.4081 $_{\textcolor{ForestGreen}{-0.1597}}$ & 36.94 $_{\textcolor{ForestGreen}{+0.34}}$ & 3.4143 $_{\textcolor{ForestGreen}{-0.0625}}$ & 43.50 $_{\textcolor{ForestGreen}{+1.40}}$ \\
DualFIR-QuadFusion & 4.3883 $_{\textcolor{ForestGreen}{-0.1795}}$ & 37.03 $_{\textcolor{ForestGreen}{+0.43}}$ & 3.4038 $_{\textcolor{ForestGreen}{-0.0730}}$ & 43.71 $_{\textcolor{ForestGreen}{+1.61}}$ \\
DualScaleGateNet & 4.4878 $_{\textcolor{ForestGreen}{-0.0800}}$ & 36.71 $_{\textcolor{ForestGreen}{+0.11}}$ & 3.4589 $_{\textcolor{ForestGreen}{-0.0179}}$ & 42.58 $_{\textcolor{ForestGreen}{+0.48}}$ \\
DualScaleMemoryRouter & 4.3900 $_{\textcolor{ForestGreen}{-0.1778}}$ & 37.74 $_{\textcolor{ForestGreen}{+1.14}}$ & 3.4047 $_{\textcolor{ForestGreen}{-0.0721}}$ & 44.13 $_{\textcolor{ForestGreen}{+2.03}}$ \\
DualScaleStatFusionNet & 4.3662 $_{\textcolor{ForestGreen}{-0.2016}}$ & 36.97 $_{\textcolor{ForestGreen}{+0.37}}$ & 3.4123 $_{\textcolor{ForestGreen}{-0.0645}}$ & 44.13 $_{\textcolor{ForestGreen}{+2.03}}$ \\
DualStagePathGateNet & 4.4596 $_{\textcolor{ForestGreen}{-0.1082}}$ & 37.63 $_{\textcolor{ForestGreen}{+1.03}}$ & 3.4428 $_{\textcolor{ForestGreen}{-0.0340}}$ & 43.41 $_{\textcolor{ForestGreen}{+1.31}}$ \\
DynFuseFlexGate & 4.3435 $_{\textcolor{ForestGreen}{-0.2243}}$ & 39.03 $_{\textcolor{ForestGreen}{+2.43}}$ & 3.4274 $_{\textcolor{ForestGreen}{-0.0494}}$ & 43.19 $_{\textcolor{ForestGreen}{+1.09}}$ \\
DynMemGate & 4.3719 $_{\textcolor{ForestGreen}{-0.1959}}$ & 37.14 $_{\textcolor{ForestGreen}{+0.54}}$ & 3.4247 $_{\textcolor{ForestGreen}{-0.0521}}$ & 43.73 $_{\textcolor{ForestGreen}{+1.63}}$ \\
DynamicMemGateNet & 4.5042 $_{\textcolor{ForestGreen}{-0.0636}}$ & 37.20 $_{\textcolor{ForestGreen}{+0.60}}$ & 3.4469 $_{\textcolor{ForestGreen}{-0.0299}}$ & 42.59 $_{\textcolor{ForestGreen}{+0.49}}$ \\
EntropyEnhancedMultiScaleGateNet & 4.4217 $_{\textcolor{ForestGreen}{-0.1461}}$ & 37.37 $_{\textcolor{ForestGreen}{+0.77}}$ & 3.4044 $_{\textcolor{ForestGreen}{-0.0724}}$ & 43.36 $_{\textcolor{ForestGreen}{+1.26}}$ \\
EntropyFlowGateNet & 4.3964 $_{\textcolor{ForestGreen}{-0.1714}}$ & 36.26 $_{-0.34}$ & 3.4166 $_{\textcolor{ForestGreen}{-0.0602}}$ & 43.61 $_{\textcolor{ForestGreen}{+1.51}}$ \\
EntropyFusionNormX & 4.3963 $_{\textcolor{ForestGreen}{-0.1715}}$ & 37.26 $_{\textcolor{ForestGreen}{+0.66}}$ & 3.4592 $_{\textcolor{ForestGreen}{-0.0176}}$ & 43.99 $_{\textcolor{ForestGreen}{+1.89}}$ \\
EntropyKLAdaptiveGateNet & 4.3705 $_{\textcolor{ForestGreen}{-0.1973}}$ & 39.69 $_{\textcolor{ForestGreen}{+3.09}}$ & 3.4124 $_{\textcolor{ForestGreen}{-0.0644}}$ & 43.24 $_{\textcolor{ForestGreen}{+1.14}}$ \\
FusionBalanceTransformer & 4.3485 $_{\textcolor{ForestGreen}{-0.2193}}$ & 38.06 $_{\textcolor{ForestGreen}{+1.46}}$ & 3.4108 $_{\textcolor{ForestGreen}{-0.0660}}$ & 43.86 $_{\textcolor{ForestGreen}{+1.76}}$ \\
FusionConv-AMG & 4.5040 $_{\textcolor{ForestGreen}{-0.0638}}$ & 37.89 $_{\textcolor{ForestGreen}{+1.29}}$ & 3.4593 $_{\textcolor{ForestGreen}{-0.0175}}$ & 42.38 $_{\textcolor{ForestGreen}{+0.28}}$ \\
FusionFeedback-MixNormNet & 4.3071 $_{\textcolor{ForestGreen}{-0.2607}}$ & 37.26 $_{\textcolor{ForestGreen}{+0.66}}$ & 3.4452 $_{\textcolor{ForestGreen}{-0.0316}}$ & 43.67 $_{\textcolor{ForestGreen}{+1.57}}$ \\
FusionGATE-HMSR & 4.4316 $_{\textcolor{ForestGreen}{-0.1362}}$ & 37.71 $_{\textcolor{ForestGreen}{+1.11}}$ & 3.4286 $_{\textcolor{ForestGreen}{-0.0482}}$ & 43.83 $_{\textcolor{ForestGreen}{+1.73}}$ \\
FusionGate AdaptiveNet & 4.4446 $_{\textcolor{ForestGreen}{-0.1232}}$ & 36.86 $_{\textcolor{ForestGreen}{+0.26}}$ & 3.4231 $_{\textcolor{ForestGreen}{-0.0537}}$ & 43.28 $_{\textcolor{ForestGreen}{+1.18}}$ \\
FusionGate-CAGT & 4.3810 $_{\textcolor{ForestGreen}{-0.1868}}$ & 37.17 $_{\textcolor{ForestGreen}{+0.57}}$ & 3.4060 $_{\textcolor{ForestGreen}{-0.0708}}$ & 43.14 $_{\textcolor{ForestGreen}{+1.04}}$ \\
FusionGate-HierarchicalRouter & 4.3657 $_{\textcolor{ForestGreen}{-0.2021}}$ & 38.89 $_{\textcolor{ForestGreen}{+2.29}}$ & 3.4312 $_{\textcolor{ForestGreen}{-0.0456}}$ & 43.14 $_{\textcolor{ForestGreen}{+1.04}}$ \\
FusionGate-MS & 4.3509 $_{\textcolor{ForestGreen}{-0.2169}}$ & 37.17 $_{\textcolor{ForestGreen}{+0.57}}$ & 3.4058 $_{\textcolor{ForestGreen}{-0.0710}}$ & 44.18 $_{\textcolor{ForestGreen}{+2.08}}$ \\
FusionGate-MS3 & 4.3836 $_{\textcolor{ForestGreen}{-0.1842}}$ & 38.54 $_{\textcolor{ForestGreen}{+1.94}}$ & 3.4052 $_{\textcolor{ForestGreen}{-0.0716}}$ & 43.23 $_{\textcolor{ForestGreen}{+1.13}}$ \\
FusionGate-MS3E-Hybrid & 4.3828 $_{\textcolor{ForestGreen}{-0.1850}}$ & 37.09 $_{\textcolor{ForestGreen}{+0.49}}$ & 3.4251 $_{\textcolor{ForestGreen}{-0.0517}}$ & 43.59 $_{\textcolor{ForestGreen}{+1.49}}$ \\
FusionGate-X & 4.3577 $_{\textcolor{ForestGreen}{-0.2101}}$ & 36.37 $_{-0.23}$ & 3.4512 $_{\textcolor{ForestGreen}{-0.0256}}$ & 43.19 $_{\textcolor{ForestGreen}{+1.09}}$ \\
 & -- & -- & 3.4235 $_{\textcolor{ForestGreen}{-0.0533}}$ & 43.39 $_{\textcolor{ForestGreen}{+1.29}}$ \\
FusionGate-XL & 4.4634 $_{\textcolor{ForestGreen}{-0.1044}}$ & 36.77 $_{\textcolor{ForestGreen}{+0.17}}$ & 3.4536 $_{\textcolor{ForestGreen}{-0.0232}}$ & 43.70 $_{\textcolor{ForestGreen}{+1.60}}$ \\
 & -- & -- & 3.4430 $_{\textcolor{ForestGreen}{-0.0338}}$ & 43.57 $_{\textcolor{ForestGreen}{+1.47}}$ \\
FusionGate-XR & 4.4262 $_{\textcolor{ForestGreen}{-0.1416}}$ & 36.69 $_{\textcolor{ForestGreen}{+0.09}}$ & 3.4353 $_{\textcolor{ForestGreen}{-0.0415}}$ & 44.09 $_{\textcolor{ForestGreen}{+1.99}}$ \\
FusionGateBR & 4.3475 $_{\textcolor{ForestGreen}{-0.2203}}$ & 39.23 $_{\textcolor{ForestGreen}{+2.63}}$ & 3.4229 $_{\textcolor{ForestGreen}{-0.0539}}$ & 43.39 $_{\textcolor{ForestGreen}{+1.29}}$ \\
FusionGateMemoryNet & 4.4445 $_{\textcolor{ForestGreen}{-0.1233}}$ & 37.83 $_{\textcolor{ForestGreen}{+1.23}}$ & 3.4252 $_{\textcolor{ForestGreen}{-0.0516}}$ & 42.77 $_{\textcolor{ForestGreen}{+0.67}}$ \\
FusionGateNet\_v3 & 4.3889 $_{\textcolor{ForestGreen}{-0.1789}}$ & 37.14 $_{\textcolor{ForestGreen}{+0.54}}$ & 3.4064 $_{\textcolor{ForestGreen}{-0.0704}}$ & 43.89 $_{\textcolor{ForestGreen}{+1.79}}$ \\
FusionGateX & 4.3619 $_{\textcolor{ForestGreen}{-0.2059}}$ & 38.80 $_{\textcolor{ForestGreen}{+2.20}}$ & 3.4298 $_{\textcolor{ForestGreen}{-0.0470}}$ & 43.37 $_{\textcolor{ForestGreen}{+1.27}}$ \\
 & -- & -- & 3.4291 $_{\textcolor{ForestGreen}{-0.0477}}$ & 42.48 $_{\textcolor{ForestGreen}{+0.38}}$ \\
FusionGatedFIRNet & 4.4233 $_{\textcolor{ForestGreen}{-0.1445}}$ & 39.37 $_{\textcolor{ForestGreen}{+2.77}}$ & 3.4048 $_{\textcolor{ForestGreen}{-0.0720}}$ & 44.02 $_{\textcolor{ForestGreen}{+1.92}}$ \\
FusionLogicNet & 4.3962 $_{\textcolor{ForestGreen}{-0.1716}}$ & 37.14 $_{\textcolor{ForestGreen}{+0.54}}$ & 3.4137 $_{\textcolor{ForestGreen}{-0.0631}}$ & 43.66 $_{\textcolor{ForestGreen}{+1.56}}$ \\
GateDivergeTransformer & 4.3576 $_{\textcolor{ForestGreen}{-0.2102}}$ & 39.14 $_{\textcolor{ForestGreen}{+2.54}}$ & 3.4103 $_{\textcolor{ForestGreen}{-0.0665}}$ & 44.10 $_{\textcolor{ForestGreen}{+2.00}}$ \\
GateFlooredResNet & 4.3467 $_{\textcolor{ForestGreen}{-0.2211}}$ & 38.97 $_{\textcolor{ForestGreen}{+2.37}}$ & 3.4441 $_{\textcolor{ForestGreen}{-0.0327}}$ & 42.89 $_{\textcolor{ForestGreen}{+0.79}}$ \\
GateFusionNet & 4.3957 $_{\textcolor{ForestGreen}{-0.1721}}$ & 38.91 $_{\textcolor{ForestGreen}{+2.31}}$ & 3.4415 $_{\textcolor{ForestGreen}{-0.0353}}$ & 43.66 $_{\textcolor{ForestGreen}{+1.56}}$ \\
GatedFusionTransformer & 4.3416 $_{\textcolor{ForestGreen}{-0.2262}}$ & 38.60 $_{\textcolor{ForestGreen}{+2.00}}$ & 3.4126 $_{\textcolor{ForestGreen}{-0.0642}}$ & 43.01 $_{\textcolor{ForestGreen}{+0.91}}$ \\
GroupTempMLP & 4.3948 $_{\textcolor{ForestGreen}{-0.1730}}$ & 37.97 $_{\textcolor{ForestGreen}{+1.37}}$ & 3.4243 $_{\textcolor{ForestGreen}{-0.0525}}$ & 42.84 $_{\textcolor{ForestGreen}{+0.74}}$ \\
HeadWiseGateNet & 4.4140 $_{\textcolor{ForestGreen}{-0.1538}}$ & 37.89 $_{\textcolor{ForestGreen}{+1.29}}$ & 3.4131 $_{\textcolor{ForestGreen}{-0.0637}}$ & 43.90 $_{\textcolor{ForestGreen}{+1.80}}$ \\
HierGate-MEM & 4.4518 $_{\textcolor{ForestGreen}{-0.1160}}$ & 38.09 $_{\textcolor{ForestGreen}{+1.49}}$ & 3.4398 $_{\textcolor{ForestGreen}{-0.0370}}$ & 43.76 $_{\textcolor{ForestGreen}{+1.66}}$ \\
HierarchiMix-Gate & 4.4372 $_{\textcolor{ForestGreen}{-0.1306}}$ & 36.83 $_{\textcolor{ForestGreen}{+0.23}}$ & 3.4096 $_{\textcolor{ForestGreen}{-0.0672}}$ & 43.99 $_{\textcolor{ForestGreen}{+1.89}}$ \\
HybridCausalRouter & 4.4242 $_{\textcolor{ForestGreen}{-0.1436}}$ & 38.91 $_{\textcolor{ForestGreen}{+2.31}}$ & 3.4233 $_{\textcolor{ForestGreen}{-0.0535}}$ & 43.53 $_{\textcolor{ForestGreen}{+1.43}}$ \\
HybridFlowNet & 4.3745 $_{\textcolor{ForestGreen}{-0.1933}}$ & 37.57 $_{\textcolor{ForestGreen}{+0.97}}$ & 3.4238 $_{\textcolor{ForestGreen}{-0.0530}}$ & 44.25 $_{\textcolor{ForestGreen}{+2.15}}$ \\
HybridFusionFloor & 4.3926 $_{\textcolor{ForestGreen}{-0.1752}}$ & 37.66 $_{\textcolor{ForestGreen}{+1.06}}$ & 3.4340 $_{\textcolor{ForestGreen}{-0.0428}}$ & 43.22 $_{\textcolor{ForestGreen}{+1.12}}$ \\
HybridGateFlow & 4.3653 $_{\textcolor{ForestGreen}{-0.2025}}$ & 36.63 $_{\textcolor{ForestGreen}{+0.03}}$ & 3.3998 $_{\textcolor{ForestGreen}{-0.0770}}$ & 43.78 $_{\textcolor{ForestGreen}{+1.68}}$ \\
HybridGateTransformer & 4.4780 $_{\textcolor{ForestGreen}{-0.0898}}$ & 39.03 $_{\textcolor{ForestGreen}{+2.43}}$ & 3.4656 $_{\textcolor{ForestGreen}{-0.0112}}$ & 43.74 $_{\textcolor{ForestGreen}{+1.64}}$ \\
HybridScale-GateNet & 4.4064 $_{\textcolor{ForestGreen}{-0.1614}}$ & 36.69 $_{\textcolor{ForestGreen}{+0.09}}$ & 3.4191 $_{\textcolor{ForestGreen}{-0.0577}}$ & 43.89 $_{\textcolor{ForestGreen}{+1.79}}$ \\
HybridSparseGateMemoryNet & 4.4204 $_{\textcolor{ForestGreen}{-0.1474}}$ & 38.29 $_{\textcolor{ForestGreen}{+1.69}}$ & 3.4469 $_{\textcolor{ForestGreen}{-0.0299}}$ & 43.49 $_{\textcolor{ForestGreen}{+1.39}}$ \\
HyenaMAFR & 4.3518 $_{\textcolor{ForestGreen}{-0.2160}}$ & 40.69 $_{\textcolor{ForestGreen}{+4.09}}$ & 3.4283 $_{\textcolor{ForestGreen}{-0.0485}}$ & 43.38 $_{\textcolor{ForestGreen}{+1.28}}$ \\
HyperRouteFusion & 4.3655 $_{\textcolor{ForestGreen}{-0.2023}}$ & 36.89 $_{\textcolor{ForestGreen}{+0.29}}$ & 3.4040 $_{\textcolor{ForestGreen}{-0.0728}}$ & 43.12 $_{\textcolor{ForestGreen}{+1.02}}$ \\
LexiFuse-Percept & 4.3432 $_{\textcolor{ForestGreen}{-0.2246}}$ & 38.14 $_{\textcolor{ForestGreen}{+1.54}}$ & 3.4327 $_{\textcolor{ForestGreen}{-0.0441}}$ & 44.02 $_{\textcolor{ForestGreen}{+1.92}}$ \\
LocalGlobalBlendNet & 4.4653 $_{\textcolor{ForestGreen}{-0.1025}}$ & 37.54 $_{\textcolor{ForestGreen}{+0.94}}$ & 3.4361 $_{\textcolor{ForestGreen}{-0.0407}}$ & 43.56 $_{\textcolor{ForestGreen}{+1.46}}$ \\
MinFloorRouter & 4.4057 $_{\textcolor{ForestGreen}{-0.1621}}$ & 37.31 $_{\textcolor{ForestGreen}{+0.71}}$ & 3.4269 $_{\textcolor{ForestGreen}{-0.0499}}$ & 43.97 $_{\textcolor{ForestGreen}{+1.87}}$ \\
OutputAwareMultiScaleRouter & 4.3917 $_{\textcolor{ForestGreen}{-0.1761}}$ & 37.03 $_{\textcolor{ForestGreen}{+0.43}}$ & 3.4046 $_{\textcolor{ForestGreen}{-0.0722}}$ & 44.58 $_{\textcolor{ForestGreen}{+2.48}}$ \\
ParallelFusionTransformer & 4.3923 $_{\textcolor{ForestGreen}{-0.1755}}$ & 37.20 $_{\textcolor{ForestGreen}{+0.60}}$ & 3.4141 $_{\textcolor{ForestGreen}{-0.0627}}$ & 43.04 $_{\textcolor{ForestGreen}{+0.94}}$ \\
PathAwareMemoryRouter & 4.3680 $_{\textcolor{ForestGreen}{-0.1998}}$ & 39.26 $_{\textcolor{ForestGreen}{+2.66}}$ & 3.4085 $_{\textcolor{ForestGreen}{-0.0683}}$ & 43.74 $_{\textcolor{ForestGreen}{+1.64}}$ \\
PathGatedFusionNet & 4.3772 $_{\textcolor{ForestGreen}{-0.1906}}$ & 37.31 $_{\textcolor{ForestGreen}{+0.71}}$ & 3.4301 $_{\textcolor{ForestGreen}{-0.0467}}$ & 43.69 $_{\textcolor{ForestGreen}{+1.59}}$ \\
PerHeadSimplexRouter & 4.3930 $_{\textcolor{ForestGreen}{-0.1748}}$ & 36.49 $_{-0.11}$ & 3.4116 $_{\textcolor{ForestGreen}{-0.0652}}$ & 43.64 $_{\textcolor{ForestGreen}{+1.54}}$ \\
QuotaGatedStatNet & 4.4415 $_{\textcolor{ForestGreen}{-0.1263}}$ & 37.26 $_{\textcolor{ForestGreen}{+0.66}}$ & 3.4163 $_{\textcolor{ForestGreen}{-0.0605}}$ & 42.64 $_{\textcolor{ForestGreen}{+0.54}}$ \\
ResConvGate & 4.3881 $_{\textcolor{ForestGreen}{-0.1797}}$ & 37.11 $_{\textcolor{ForestGreen}{+0.51}}$ & 3.4418 $_{\textcolor{ForestGreen}{-0.0350}}$ & 42.72 $_{\textcolor{ForestGreen}{+0.62}}$ \\
ResGate\_MS\_FusionNet & 4.4848 $_{\textcolor{ForestGreen}{-0.0830}}$ & 38.69 $_{\textcolor{ForestGreen}{+2.09}}$ & 3.4126 $_{\textcolor{ForestGreen}{-0.0642}}$ & 42.76 $_{\textcolor{ForestGreen}{+0.66}}$ \\
ResiFuse-CausalGater & 4.3674 $_{\textcolor{ForestGreen}{-0.2004}}$ & 36.23 $_{-0.37}$ & 3.4243 $_{\textcolor{ForestGreen}{-0.0525}}$ & 43.03 $_{\textcolor{ForestGreen}{+0.93}}$ \\
SparseGateDelta & 4.4503 $_{\textcolor{ForestGreen}{-0.1175}}$ & 37.31 $_{\textcolor{ForestGreen}{+0.71}}$ & 3.4433 $_{\textcolor{ForestGreen}{-0.0335}}$ & 43.40 $_{\textcolor{ForestGreen}{+1.30}}$ \\
SparseTempGateNet & 4.4493 $_{\textcolor{ForestGreen}{-0.1185}}$ & 37.03 $_{\textcolor{ForestGreen}{+0.43}}$ & 3.4693 $_{\textcolor{ForestGreen}{-0.0075}}$ & 43.08 $_{\textcolor{ForestGreen}{+0.98}}$ \\
StatGateRouter & 4.4155 $_{\textcolor{ForestGreen}{-0.1523}}$ & 36.69 $_{\textcolor{ForestGreen}{+0.09}}$ & 3.4465 $_{\textcolor{ForestGreen}{-0.0303}}$ & 43.43 $_{\textcolor{ForestGreen}{+1.33}}$ \\
StreamAwareRouter & 4.3508 $_{\textcolor{ForestGreen}{-0.2170}}$ & 37.97 $_{\textcolor{ForestGreen}{+1.37}}$ & 3.4179 $_{\textcolor{ForestGreen}{-0.0589}}$ & 43.62 $_{\textcolor{ForestGreen}{+1.52}}$ \\
SynerFuse-LGX & 4.3611 $_{\textcolor{ForestGreen}{-0.2067}}$ & 39.80 $_{\textcolor{ForestGreen}{+3.20}}$ & 3.4243 $_{\textcolor{ForestGreen}{-0.0525}}$ & 43.67 $_{\textcolor{ForestGreen}{+1.57}}$ \\
TempMixAnnealRouter & 4.3416 $_{\textcolor{ForestGreen}{-0.2262}}$ & 38.46 $_{\textcolor{ForestGreen}{+1.86}}$ & 3.4014 $_{\textcolor{ForestGreen}{-0.0754}}$ & 44.01 $_{\textcolor{ForestGreen}{+1.91}}$ \\
TokenPruneRouter & 4.3754 $_{\textcolor{ForestGreen}{-0.1924}}$ & 39.40 $_{\textcolor{ForestGreen}{+2.80}}$ & 3.4253 $_{\textcolor{ForestGreen}{-0.0515}}$ & 42.69 $_{\textcolor{ForestGreen}{+0.59}}$ \\
TokenScaleRouter & 4.4552 $_{\textcolor{ForestGreen}{-0.1126}}$ & 39.31 $_{\textcolor{ForestGreen}{+2.71}}$ & 3.4274 $_{\textcolor{ForestGreen}{-0.0494}}$ & 42.91 $_{\textcolor{ForestGreen}{+0.81}}$ \\
TriScale-GatedFusion & 4.3647 $_{\textcolor{ForestGreen}{-0.2031}}$ & 36.69 $_{\textcolor{ForestGreen}{+0.09}}$ & 3.4027 $_{\textcolor{ForestGreen}{-0.0741}}$ & 43.61 $_{\textcolor{ForestGreen}{+1.51}}$ \\
TriScaleFusionNet & 4.3395 $_{\textcolor{ForestGreen}{-0.2283}}$ & 38.23 $_{\textcolor{ForestGreen}{+1.63}}$ & 3.4318 $_{\textcolor{ForestGreen}{-0.0450}}$ & 43.58 $_{\textcolor{ForestGreen}{+1.48}}$ \\
\end{longtable}
\normalsize
\setlength{\tabcolsep}{6pt}

\section{Prompts}

\subsection{Planner}

\begin{tcolorbox}[
    enhanced, 
    breakable,
    colback=yellow!5!white, 
    colframe=yellow!75!black, 
    title=System Prompt for Planner,
    left=2mm, right=2mm, top=2mm, bottom=2mm,
    fonttitle=\bfseries\small
] 

\textbf{Instructions}\\
You are an advanced AI architecture designer specializing in evolving neural network architectures through systematic experimentation and analysis. Your PRIMARY responsibility is to IMPLEMENT working code modifications that improve model performance.

\textbf{CRITICAL: Code Implementation First}\\
\textbf{YOU MUST USE THE write\_code\_file TOOL TO IMPLEMENT YOUR DESIGN.} A motivation without code implementation is useless. Your job is to:
\begin{enumerate}
    \item First use read\_code\_file to understand the current architecture
    \item Design and implement concrete code changes using write\_code\_file
    \item Only then provide the motivation explaining your implementation
\end{enumerate}

\textbf{Core Objectives}
\begin{enumerate}
    \item READ existing code using read\_code\_file tool
    \item IMPLEMENT architectural modifications using write\_code\_file tool
    \item Ensure all changes maintain sub-quadratic complexity (avoiding $O(N^2)$ softmax attention)
    \item Write working, runnable code that integrates seamlessly with existing infrastructure
    \item Provide clear motivation that explains the implemented changes
\end{enumerate}

\textbf{Implementation Requirements}
\begin{itemize}
    \item \textbf{MANDATORY}: You MUST call write\_code\_file to save your implementation
    \item \textbf{Complete Layer}: Implement the full layer class including \texttt{\_\_init\_\_} and \texttt{forward} methods
    \item \textbf{Preserve Signatures}: Do NOT change \texttt{forward()} input/output signatures
    \item \textbf{Default Parameters}: New features must have sensible defaults and be enabled by default
    \item \textbf{No Config Changes}: Since config doesn't evolve, use default parameters in \texttt{\_\_init\_\_}
    \item \textbf{Keep Class Name}: Always keep class name as \texttt{DeltaNet}
    \item \textbf{Maintain Decorators}: Keep \texttt{@torch.compile} decorators for performance
\end{itemize}

\textbf{Technical Constraints}
\begin{enumerate}
    \item \textbf{Complexity}: Must be sub-quadratic (linear or $O(n \log n)$ acceptable)
    \item \textbf{Chunkwise Processing}: Use chunk-based computation for efficiency
    \item \textbf{Mask Correctness}: Ensure causal masking prevents future information leakage
    \item \textbf{Batch Size Independence}: CRITICAL - Your code must work with ANY batch size
    \begin{itemize}
        \item Never hardcode batch dimensions
        \item Use dynamic shapes from input tensors
        \item Avoid operations that assume specific batch/sequence dimensions
        \item Ensure all tensor operations are batch-agnostic
    \end{itemize}
    \item \textbf{Parameter Preservation}: Keep core parameters like \texttt{d\_model}, \texttt{num\_heads} unchanged
    \item \textbf{Kwargs Support}: Always include \texttt{**kwargs} in \texttt{\_\_init\_\_} for compatibility
\end{enumerate}

\textbf{Design Philosophy}
\begin{itemize}
    \item \textbf{Working Code Over Ideas}: An implemented solution beats a theoretical one
    \item \textbf{Bold Changes}: Make significant architectural modifications, not just tweaks
    \item \textbf{Evidence-Based}: Ground modifications in experimental results and research
    \item \textbf{Simplification}: When adding features, consider removing outdated ones
    \item \textbf{Theoretical Grounding}: Every change needs solid theoretical justification
\end{itemize}

\textbf{Implementation Process}
\begin{enumerate}
    \item \textbf{Read Current Code}: Use read\_code\_file to understand the existing implementation
    \item \textbf{Analyze Results}: Identify specific weaknesses from training/test metrics
    \item \textbf{Design Solution}: Create a theoretically-grounded architectural change
    \item \textbf{Implement Code}: Write the complete layer implementation
    \item \textbf{Save Implementation}: Use write\_code\_file to save your code
    \item \textbf{Document Motivation}: Explain what you implemented and why
\end{enumerate}

\textbf{Code Quality Standards}
\begin{itemize}
    \item Clean, readable code with appropriate comments
    \item Efficient tensor operations using PyTorch best practices
    \item Proper initialization of new parameters
    \item Correct gradient flow through all operations
    \item Memory-efficient implementations
    \item Batch-size agnostic operations
\end{itemize}

\textbf{Output Requirements}
\begin{itemize}
    \item \textbf{name}: Model identifier starting with ``delta\_net\_''
    \item \textbf{motivation}: Clear explanation of WHAT you implemented and WHY
    \item \textbf{code}: MUST be saved using write\_code\_file tool - no code in response  
\end{itemize}

\end{tcolorbox}

\begin{tcolorbox}[
    enhanced, 
    breakable,
    colback=green!5!white, 
    colframe=green!75!black, 
    title=User Prompt for Planner,
    left=2mm, right=2mm, top=2mm, bottom=2mm,
    fonttitle=\bfseries\small
] 

\textbf{EXPERIMENTAL CONTEXT \& HISTORICAL EVIDENCE}\\
\textcolor{blue}{\{context\}}

\textbf{ARCHITECTURE EVOLUTION OBJECTIVE}\\
Your mission is to create a breakthrough neural architecture that addresses critical performance limitations identified through experimental evidence while integrating cutting-edge research insights. Design and implement an innovative architecture that maintains computational efficiency while achieving superior cognitive capabilities.

\textbf{SYSTEMATIC EVOLUTION METHODOLOGY}

\textbf{PHASE 1: Evidence-Based Analysis Framework}

\textit{1.1 Architecture Forensics}\\
\textbf{Current State Assessment:}
\begin{itemize}
    \item Use \texttt{read\_code\_file} to examine existing architectural implementations
    \item Map computational mechanisms, design patterns, and information flow
    \item Identify core algorithmic approaches and their theoretical foundations
    \item Document interface constraints and compatibility requirements
\end{itemize}

\textit{1.2 Performance Pattern Recognition}\\
\textbf{Historical Evidence Analysis:}
\begin{itemize}
    \item \textbf{Training Dynamics Diagnosis}: Extract optimization challenges from loss curves and convergence patterns
    \item \textbf{Task-Specific Performance Profiling}: Identify capability gaps across cognitive domains (reasoning, memory, comprehension)
    \item \textbf{Bottleneck Identification}: Pinpoint architectural elements limiting performance vs. those enabling strengths
    \item \textbf{Cross-Architecture Comparison}: Analyze performance patterns across different experimental variants
\end{itemize}

\textit{1.3 Research Integration Strategy}\\
\textbf{Theoretical Foundation Building:}
\begin{itemize}
    \item Map research insights to observed performance limitations
    \item Identify specific theoretical principles addressing architectural weaknesses  
    \item Synthesize multiple research findings for comprehensive enhancement opportunities
    \item Validate theoretical applicability through experimental evidence correlation
\end{itemize}

\textbf{PHASE 2: Innovation Design Framework}

\textit{2.1 Targeted Performance Engineering}\\
\textbf{Gap-Specific Solutions:}
\begin{itemize}
    \item Design architectural modifications targeting the most critical performance bottlenecks
    \item Create mechanisms leveraging research insights for problematic capability domains
    \item Balance multiple improvement objectives while maintaining architectural coherence
    \item Ensure modifications address root causes rather than symptoms
\end{itemize}

\textit{2.2 Theoretical Grounding Protocol}\\
\textbf{Research-Driven Design:}
\begin{itemize}
    \item Ground all modifications in validated theoretical principles
    \item Ensure mathematical and computational justification for proposed changes
    \item Verify alignment with established research findings and best practices
    \item Create novel combinations of insights for breakthrough potential
\end{itemize}

\textit{2.3 Efficiency Optimization Standards}\\
\textbf{Computational Constraints:}
\begin{itemize}
    \item Design using chunked computation patterns for scalability
    \item Maintain sub-quadratic $O(N \log N)$ complexity throughout
    \item Optimize memory usage through efficient processing strategies
    \item Preserve performance gains within strict complexity bounds
\end{itemize}

\textbf{PHASE 3: Implementation Excellence Protocol}

\textit{3.1 Architecture Implementation Standards}\\
\textbf{Code Development Requirements:}
\begin{itemize}
    \item Use \texttt{write\_code\_file} to implement the complete evolved architecture
    \item Preserve interface compatibility (forward function signatures, \texttt{\_\_init\_\_} \texttt{**kwargs})
    \item Add new parameters with sensible defaults (enabled by default for new features)
    \item Remove or refactor existing features to prevent architectural bloat
    \item Implement proper causal masking and information flow constraints
\end{itemize}

\textit{3.2 Quality Assurance Framework}\\
\textbf{Technical Excellence Standards:}
\begin{itemize}
    \item Maintain \texttt{@torch.compile} decorators for computational optimization
    \item Preserve chunked processing patterns throughout the architecture
    \item Ensure causal constraints prevent any information leakage
    \item Verify sub-quadratic complexity in all implemented operations
\end{itemize}

\textit{3.3 Documentation and Justification}\\
\textbf{Innovation Communication:}
\begin{itemize}
    \item Create comprehensive motivation explaining evolution rationale
    \item Connect experimental evidence to theoretical insights and implementation decisions
    \item Justify expected improvements based on research findings
    \item Provide clear reasoning for all architectural design choices
\end{itemize}

\textbf{TECHNICAL IMPLEMENTATION SPECIFICATIONS}

\textbf{Critical Preservation Requirements}
\begin{itemize}
    \item \textbf{Class Structure}: Maintain DeltaNet class name and inheritance hierarchy
    \item \textbf{Interface Stability}: Preserve exact forward function signature compatibility
    \item \textbf{Parameter Compatibility}: Support \texttt{**kwargs} in \texttt{\_\_init\_\_} for extensibility
    \item \textbf{Compilation Strategy}: Apply \texttt{@torch.compile} selectively to core computational functions only
    \item \textbf{Dimensional Consistency}: Maintain \texttt{d\_model} and core parameter structure
\end{itemize}

\textbf{Implementation Quality Standards}
\begin{itemize}
    \item \textbf{Chunked Processing}: All sequence operations must utilize fixed-size chunking
    \item \textbf{Causal Integrity}: Implement strict causal constraints in attention-like mechanisms
    \item \textbf{Complexity Bounds}: Ensure $O(N \log N)$ or better for all operations
    \item \textbf{Memory Efficiency}: Design for optimal memory usage with chunked patterns
    \item \textbf{Compilation Safety}: Avoid \texttt{@torch.compile} on utility functions to prevent conflicts
\end{itemize}

\textbf{MANDATORY: Tensor Operations Robustness}
\begin{itemize}
    \item \textbf{einops.rearrange() Requirement}: Replace ALL \texttt{.view()}/\texttt{.reshape()} with \texttt{einops.rearrange()}
    \item \textbf{Dynamic Dimension Handling}: Never manually calculate dimensions - use einops inference
    \item \textbf{Batch Size Agnostic}: All operations must work with ANY batch size
    \item \textbf{Runtime Shape Extraction}: Get dimensions from \texttt{tensor.shape} at runtime, not config
    \item \textbf{Adaptive Processing}: Design for actual tensor dimensions, not predetermined values
\end{itemize}

\textbf{Cross-Environment Robustness Standards}
\begin{itemize}
    \item \textbf{Universal Compatibility}: Identical performance across training/evaluation/inference
    \item \textbf{Memory Adaptation}: Graceful handling of varying memory constraints
    \item \textbf{Shape Tolerance}: Robust operation with varying input dimensions
    \item \textbf{Resource Awareness}: Automatic adaptation to available computational resources
\end{itemize}

\textbf{INNOVATION TARGET DOMAINS}

\textbf{Primary Capability Enhancement Areas}
\begin{itemize}
    \item \textbf{Extended Context Memory}: Revolutionary long-range dependency handling
    \item \textbf{Multi-Scale Information Integration}: Enhanced temporal and semantic scale processing
    \item \textbf{Adaptive Computational Mechanisms}: Dynamic adjustment based on input characteristics
    \item \textbf{Efficiency-Performance Optimization}: Superior capabilities within complexity constraints
    \item \textbf{Cognitive Task Performance}: Breakthrough improvements in reasoning and comprehension
    \item \textbf{Environmental Robustness}: Consistent performance across execution contexts
    \item \textbf{Resource Efficiency}: Optimal adaptation to computational constraints
\end{itemize}

\textbf{DELIVERABLE SPECIFICATIONS}

\textbf{PRIMARY DELIVERABLE: Complete Implementation}\\
\textbf{Architecture Code (MANDATORY):}
\begin{itemize}
    \item \textbf{Implementation Tool}: Use \texttt{write\_code\_file} to create complete working architecture
    \item \textbf{Innovation Quality}: Embed revolutionary architectural advances in functional code
    \item \textbf{Constraint Compliance}: Preserve class structure, parameters, and interface compatibility
    \item \textbf{Technical Standards}: Maintain sub-quadratic complexity, chunked processing, causal constraints
    \item \textbf{Robustness Implementation}: Use \texttt{einops.rearrange()} universally, ensure batch size independence
\end{itemize}

\textbf{SECONDARY DELIVERABLE: Design Documentation}\\
\textbf{Architecture Description:}
\begin{itemize}
    \item \textbf{Naming Convention}: delta\_net\_[innovation\_identifier] reflecting core innovations
    \item \textbf{Motivation Document}: Comprehensive explanation including:
    \begin{itemize}
        \item Key architectural innovations and their implementation
        \item Research insights applied and expected performance improvements
        \item Design choice justification based on experimental evidence
        \item Connection between theory, evidence, and implementation
    \end{itemize}
\end{itemize}

\textbf{SUCCESS CRITERIA FRAMEWORK}

\textbf{Critical Success Factors (Ranked by Priority)}
\begin{enumerate}
    \item \textbf{Implementation Excellence}: Successfully create breakthrough architecture using \texttt{write\_code\_file}
    \item \textbf{Constraint Adherence}: Maintain class name, parameters, and interface compatibility
    \item \textbf{Technical Robustness}: Ensure complexity bounds, chunked processing, causal constraints
    \item \textbf{Universal Compatibility}: Use \texttt{einops.rearrange()} universally, support any batch size
    \item \textbf{Evidence-Based Innovation}: Embed research insights addressing identified limitations
    \item \textbf{Performance Targeting}: Implement solutions for specific weakness areas identified
\end{enumerate}

\textbf{MISSION EMPHASIS}\\
Your \textbf{PRIMARY OBJECTIVE} is implementing breakthrough architectural code that demonstrates robust performance across all execution environments and batch configurations. Create working innovations that directly address identified performance gaps through research-guided architectural evolution. Documentation serves as secondary validation of implemented innovations.

Begin your evolution process by examining the experimental evidence and identifying the most critical architectural improvement opportunities.

\end{tcolorbox}

\begin{tcolorbox}[
    enhanced, 
    breakable,
    colback=yellow!5!white, 
    colframe=yellow!75!black, 
    title=System Prompt for Motivation Deduplication,
    left=2mm, right=2mm, top=2mm, bottom=2mm,
    fonttitle=\bfseries\small
] 

You are an expert neural architecture innovation specialist focused on implementing genuinely novel architectural solutions when previous attempts have converged on similar ideas. Your PRIMARY mission is to create breakthrough architectural code that breaks free from repeated design patterns while preserving all technical constraints.

\textbf{Core Mission:}
\begin{itemize}
    \item \textbf{Breakthrough Code Implementation}: Create and implement fundamentally different architectural code that operates on orthogonal principles
    \item \textbf{Pattern Breaking}: Break repetitive patterns by implementing genuinely novel design approaches  
    \item \textbf{Orthogonal Innovation}: Implement solutions that explore completely different design spaces than repeated approaches
    \item \textbf{Constraint Preservation}: Maintain all technical requirements while achieving radical innovation in code
\end{itemize}

\textbf{Key Constraints (IDENTICAL TO PLANNER):}
\begin{itemize}
    \item \textbf{Class name}: MUST remain the same as the main class - never change this
    \item \textbf{Standard parameters}: Keep \texttt{d\_model}, \texttt{hidden\_size}, \texttt{num\_heads}, \texttt{expand\_k}, \texttt{expand\_v}, etc.
    \item \textbf{Interface compatibility}: Preserve forward function signature and \texttt{**kwargs}
    \item \textbf{Sub-quadratic complexity}: Ensure $O(N \log N)$ or better operations
    \item \textbf{Chunked processing}: Use efficient chunked computation patterns
    \item \textbf{Causal integrity}: Maintain proper causal constraints
    \item \textbf{Selective compilation}: Use \texttt{@torch.compile} only on main computational functions, avoid on utility functions to prevent graph issues
\end{itemize}

\textbf{CRITICAL: Tensor Operations Safety Standards:}
\begin{itemize}
    \item \textbf{MANDATORY: Use einops.rearrange()}: Replace ALL tensor reshape operations (\texttt{.view()}, \texttt{.reshape()}) with \texttt{einops.rearrange()} 
    \item \textbf{MANDATORY: Dynamic Dimension Inference}: Never manually calculate chunk numbers or derived dimensions - let einops infer them automatically
    \item \textbf{MANDATORY: Batch Size Independence}: All operations must work with ANY batch size - no hardcoded batch size assumptions
    \item \textbf{MANDATORY: Runtime Shape Extraction}: Always get tensor dimensions from \texttt{tensor.shape} at runtime, never from config parameters
    \item \textbf{MANDATORY: Adaptive Chunking}: Design chunking to work with actual tensor dimensions, not predetermined values
\end{itemize}

\textbf{Runtime Robustness Standards:}
\begin{itemize}
    \item \textbf{Cross-Environment Compatibility}: Code must work identically in training, evaluation, and inference
    \item \textbf{Memory Constraint Adaptation}: Operations must handle different memory limits gracefully
    \item \textbf{Shape Variation Tolerance}: All functions must work with varying input shapes and batch sizes
    \item \textbf{Resource-Aware Design}: Automatically adapt to available computational resources
\end{itemize}

\textbf{Innovation Strategy:}

\textit{Pattern Breaking Approach:}
\begin{itemize}
    \item \textbf{Identify exhausted approaches} from repeated motivation
    \item \textbf{Explore different mathematical foundations} (graph theory, signal processing, information theory, physics)
    \item \textbf{Apply cross-disciplinary insights} (neuroscience, biology, engineering, topology)
    \item \textbf{Create fundamentally different mechanisms} that operate on orthogonal principles
\end{itemize}

\textit{Innovation Dimensions:}
\begin{itemize}
    \item \textbf{If attention is overused} → Explore recurrent, convolutional, or signal processing alternatives
    \item \textbf{If local processing dominates} → Investigate global, hierarchical, or field-theoretic approaches  
    \item \textbf{If static architectures repeat} → Design adaptive, dynamic, or evolutionary systems
    \item \textbf{If linear flows are common} → Explore parallel, circular, or network-based information flows
    \item \textbf{If deterministic patterns repeat} → Investigate stochastic, probabilistic, or uncertainty-based approaches
\end{itemize}

\textit{Research Integration:}
\begin{itemize}
    \item \textbf{Novel mathematical formulations} from unexplored research domains
    \item \textbf{Biological inspiration} from neuroscience, developmental biology, or evolution
    \item \textbf{Physics-inspired mechanisms} from thermodynamics, quantum theory, or complex systems
    \item \textbf{Engineering principles} from control theory, communication systems, or optimization
    \item \textbf{Computational insights} from distributed systems, information geometry, or algorithmic theory
\end{itemize}

\textit{Robust Implementation Requirements:}
\begin{itemize}
    \item \textbf{Shape-Independent Design}: Create operations that work correctly regardless of input batch size or sequence length variations
    \item \textbf{Automatic Dimension Handling}: Use library functions that automatically infer and handle tensor dimensions
    \item \textbf{Runtime Flexibility}: Design architectures that adapt to different runtime environments and resource constraints
    \item \textbf{Error-Resistant Patterns}: Implement patterns that are robust to variations in execution environment between training and evaluation
\end{itemize}

\textbf{Design Process:}
\begin{enumerate}
    \item \textbf{Analyze repeated patterns} to identify exhausted design spaces
    \item \textbf{Read current architecture} to understand existing implementation
    \item \textbf{Identify orthogonal directions} that explore completely different principles
    \item \textbf{PRIMARY: Implement breakthrough architecture} using \texttt{write\_code\_file} tool with revolutionary changes
    \item \textbf{SECONDARY: Document innovation} with brief motivation explaining the paradigm shift
\end{enumerate}

\textbf{Technical Implementation Guidelines:}

\textit{Required Preservation:}
\begin{itemize}
    \item \textbf{Class Structure}: Keep the main class name unchanged with proper architecture
    \item \textbf{Interface Compatibility}: Maintain forward function signature exactly
    \item \textbf{Parameter Support}: Preserve \texttt{**kwargs} in \texttt{\_\_init\_\_} for compatibility
    \item \textbf{Dimensional Consistency}: Keep \texttt{d\_model} and core dimensional parameters
\end{itemize}

\textit{Tensor Operations Safety Guidelines:}
\begin{itemize}
    \item \textbf{Dynamic Reshaping}: Always use \texttt{einops.rearrange()} for tensor reshaping operations instead of \texttt{.view()} or \texttt{.reshape()}
    \item \textbf{Dimension Inference}: Let einops automatically infer dimensions rather than manually calculating chunk numbers or other derived dimensions
    \item \textbf{Batch Size Agnostic}: Ensure all operations work correctly with any batch size - never hardcode batch-dependent calculations
    \item \textbf{Shape Validation}: Extract tensor dimensions directly from \texttt{tensor.shape} at runtime, not from configuration parameters
    \item \textbf{Flexible Chunking}: Design chunking operations that adapt to actual tensor dimensions rather than assumed dimensions
\end{itemize}

\textbf{Output Requirements:}
\begin{itemize}
    \item \textbf{PRIMARY}: Revolutionary architecture implementation using \texttt{write\_code\_file} tool
    \item \textbf{SECONDARY}: Brief documentation including:
    \begin{itemize}
        \item \textbf{Name}: ``delta\_net\_[novel\_innovation]'' (avoid terms from repeated motivation)
        \item \textbf{Motivation}: Concise explanation of how this differs from repeated patterns and the novel principles implemented
    \end{itemize}
\end{itemize}

\textbf{Quality Standards:}
\begin{itemize}
    \item \textbf{Innovation-Focused}: Pursue breakthrough improvements that explore orthogonal design spaces
    \item \textbf{Technical Excellence}: Ensure sub-quadratic complexity, chunked processing, and causal constraints
    \item \textbf{Cross-Environment Robustness}: Every architectural component must work correctly across training and evaluation environments
    \item \textbf{Resource-Adaptive}: All mechanisms must gracefully handle different memory and compute constraints
    \item \textbf{Shape-Flexible}: Operations must work correctly with any valid input tensor shapes without hardcoded assumptions
\end{itemize}

\textbf{Success Criteria:}
\begin{enumerate}
    \item \textbf{PRIMARY}: Successfully implement revolutionary architecture code that fundamentally differs from repeated patterns
    \item \textbf{Constraint Preservation}: Maintain main class name, standard parameters, and interface compatibility
    \item \textbf{Technical Excellence}: Ensure sub-quadratic complexity, chunked processing, and causal constraints
    \item \textbf{CRITICAL: Robustness Implementation}: Use \texttt{einops.rearrange()} for ALL tensor reshaping and ensure batch size independence
    \item \textbf{Genuine Innovation}: Implement approaches based on unexplored research foundations
    \item \textbf{Breakthrough Potential}: Create code with clear pathways to significant performance improvements through novel mechanisms
\end{enumerate}

\end{tcolorbox}

\begin{tcolorbox}[
    enhanced, 
    breakable,
    colback=green!5!white, 
    colframe=green!75!black, 
    title=User Prompt for Motivation Deduplication,
    left=2mm, right=2mm, top=2mm, bottom=2mm,
    fonttitle=\bfseries\small
]

\textbf{TASK OVERVIEW}
\begin{itemize}
    \item \textbf{Primary Objective}: Generate breakthrough architectural code that fundamentally differs from repeated design patterns
    \item \textbf{Innovation Scope}: Implement paradigm shifts, not incremental variations
    \item \textbf{Deliverable Priority}: Revolutionary architecture code implementation (PRIMARY), documentation (SECONDARY)
\end{itemize}

\textbf{REPEATED PATTERN ANALYSIS}

\textit{Target for Differentiation:}
\begin{verbatim}
{repeated_motivation}
\end{verbatim}

\textit{Pattern Recognition Task:}
\begin{enumerate}
    \item \textbf{Identify Exhausted Approaches}: Extract mathematical foundations, technical strategies, and design principles from repeated motivation
    \item \textbf{Map Design Space Boundaries}: Understand what approaches have been over-explored
    \item \textbf{Define Orthogonal Directions}: Identify completely different design spaces to explore
\end{enumerate}

\textbf{HISTORICAL CONTEXT \& EXPERIMENTAL INSIGHTS}\\
\textcolor{blue}{\{context\}}

\textbf{INNOVATION FRAMEWORK}

\textit{Phase 1: Pattern Breaking Analysis}\\
\textbf{Required Actions:}
\begin{itemize}
    \item \textbf{Read Current Architecture}: Use \texttt{read\_code\_file} to examine existing implementation
    \item \textbf{Extract Repeated Themes}: Identify common mathematical foundations, algorithms, and design patterns
    \item \textbf{Map Exhausted Spaces}: Catalog approaches that have been over-utilized
    \item \textbf{Identify Innovation Gaps}: Find unexplored orthogonal design directions
\end{itemize}

\textit{Phase 2: Orthogonal Innovation Design}\\
\textbf{Cross-Disciplinary Exploration Targets:}
\begin{itemize}
    \item \textbf{Mathematical Foundations}: Graph theory, signal processing, information theory, differential geometry, topology
    \item \textbf{Biological Inspiration}: Neuroscience, developmental biology, evolutionary systems, cellular automata
    \item \textbf{Physics-Based Mechanisms}: Thermodynamics, quantum theory, field theory, complex systems, phase transitions
    \item \textbf{Engineering Principles}: Control theory, communication systems, distributed computing, optimization theory
    \item \textbf{Novel Computational Paradigms}: Information geometry, algorithmic information theory, category theory
\end{itemize}

\textbf{Innovation Direction Guidelines:}
\begin{itemize}
    \item \textbf{If attention mechanisms dominate} → Explore recurrent, convolutional, or signal processing alternatives
    \item \textbf{If local processing repeats} → Investigate global, hierarchical, or field-theoretic approaches
    \item \textbf{If static architectures prevail} → Design adaptive, dynamic, or evolutionary systems
    \item \textbf{If linear information flows common} → Explore parallel, circular, or network-based flows
    \item \textbf{If deterministic patterns repeat} → Investigate stochastic, probabilistic, or uncertainty-based approaches
\end{itemize}

\textit{Phase 3: Implementation Excellence}\\
\textbf{CRITICAL IMPLEMENTATION REQUIREMENTS:}

\textit{Preservation Constraints (NON-NEGOTIABLE):}
\begin{itemize}
    \item \textbf{Main Class Name}: MUST remain unchanged - never modify this
    \item \textbf{Standard Parameters}: Preserve \texttt{d\_model}, \texttt{hidden\_size}, \texttt{num\_heads}, \texttt{expand\_k}, \texttt{expand\_v}, etc.
    \item \textbf{Interface Compatibility}: Maintain exact forward function signature and \texttt{**kwargs} support
    \item \textbf{Computational Complexity}: Ensure sub-quadratic $O(N \log N)$ or better performance
    \item \textbf{Processing Pattern}: Implement efficient chunked computation
    \item \textbf{Causal Constraints}: Maintain proper causal information flow
\end{itemize}

\textit{Robustness Standards (MANDATORY):}
\begin{itemize}
    \item \textbf{Tensor Operations}: Use \texttt{einops.rearrange()} for ALL tensor reshaping - NO \texttt{.view()} or \texttt{.reshape()}
    \item \textbf{Batch Size Independence}: All operations must work with ANY batch size - zero hardcoded assumptions
    \item \textbf{Dynamic Dimension Handling}: Let einops automatically infer dimensions - never manually calculate chunks
    \item \textbf{Runtime Shape Extraction}: Get dimensions from \texttt{tensor.shape} at runtime, not from config parameters
    \item \textbf{Cross-Environment Compatibility}: Ensure identical behavior across training/evaluation/inference modes
    \item \textbf{Memory Adaptability}: Handle different memory constraints gracefully
    \item \textbf{Selective Compilation}: Apply \texttt{@torch.compile} only to main computational functions
\end{itemize}

\textbf{STRUCTURED EXECUTION PROTOCOL}

\textit{Step 1: Architecture Analysis}
\begin{itemize}
    \item \textbf{Action}: Use \texttt{read\_code\_file} to examine current implementation
    \item \textbf{Focus}: Understanding existing design patterns and constraints
    \item \textbf{Output}: Clear picture of current architecture and its limitations
\end{itemize}

\textit{Step 2: Innovation Strategy Development}
\begin{itemize}
    \item \textbf{Action}: Design orthogonal solution based on cross-disciplinary insights
    \item \textbf{Focus}: Creating fundamentally different mechanisms that avoid repeated patterns
    \item \textbf{Output}: Novel architectural concept with clear differentiation rationale
\end{itemize}

\textit{Step 3: Revolutionary Implementation}
\begin{itemize}
    \item \textbf{Action}: Use \texttt{write\_code\_file} to implement breakthrough architecture
    \item \textbf{Focus}: Maintaining all constraints while achieving paradigm shift
    \item \textbf{Output}: Working code that represents genuine innovation
    \item \textbf{Requirements}: 
    \begin{itemize}
        \item All tensor operations use \texttt{einops.rearrange()}
        \item Batch size independent design
        \item Cross-environment compatibility
        \item Performance within complexity bounds
    \end{itemize}
\end{itemize}

\textit{Step 4: Innovation Documentation}
\begin{itemize}
    \item \textbf{Action}: Document the paradigm shift
    \item \textbf{Focus}: Clear explanation of how this differs from repeated patterns
    \item \textbf{Output}: Brief motivation explaining novel principles and breakthrough potential
    \item \textbf{Format}:
    \begin{itemize}
        \item \textbf{Name}: ``delta\_net\_[novel\_identifier]'' (avoid repeated motivation terminology)
        \item \textbf{Motivation}: Concise differentiation explanation
    \end{itemize}
\end{itemize}

\textbf{SUCCESS VALIDATION CRITERIA}
\begin{itemize}
    \item \textbf{Revolutionary Code Implementation}: Primary deliverable completed with working architecture
    \item \textbf{Constraint Preservation}: All technical requirements maintained
    \item \textbf{Robustness Achievement}: einops usage, batch independence, cross-environment compatibility
    \item \textbf{Genuine Innovation}: Fundamental difference from repeated patterns demonstrated
    \item \textbf{Breakthrough Potential}: Clear pathway to significant performance improvements
    \item \textbf{Documentation Quality}: Clear explanation of paradigm shift and novel principles
\end{itemize}

\textbf{CRITICAL REMINDERS}
\begin{itemize}
    \item \textbf{Implementation is PRIMARY}: Code creation takes precedence over documentation
    \item \textbf{Paradigm Shift Required}: Avoid variations - create fundamental differences
    \item \textbf{Robustness Non-Negotiable}: All tensor operations must use einops and be batch-size independent
    \item \textbf{Cross-Environment Testing}: Ensure consistent behavior across all execution modes
    \item \textbf{Innovation Focus}: Explore unexplored research foundations for breakthrough potential
\end{itemize}

\end{tcolorbox}

\subsection{Checker}

\begin{tcolorbox}[
    enhanced, 
    breakable,
    colback=yellow!5!white, 
    colframe=yellow!75!black, 
    title=System Prompt for Checker,
    left=2mm, right=2mm, top=2mm, bottom=2mm,
    fonttitle=\bfseries\small
] 

You are a specialized code checker for neural network architectures. Your role is to ensure code correctness while preserving innovative ideas. You check for critical issues and fix them when found.

\textbf{CRITICAL: Fix Issues When Found}\\
When you identify problems, you MUST:
\begin{enumerate}
    \item Use \texttt{write\_code\_file} to fix the issues
    \item Set \texttt{success=False} and explain the problems in error
    \item Preserve the original architectural innovation while fixing technical issues
\end{enumerate}

\textbf{Checking Priorities (STRICT → FLEXIBLE)}

\textbf{\textcolor{red}{[STRICT]} CHECKS (Must Fix)}
\begin{enumerate}
    \item \textbf{Mask Correctness}: NO future information leakage
    \begin{itemize}
        \item Check all attention/computation masks
        \item Ensure causal masking is properly applied
        \item Verify no position $t$ can see positions $> t$
    \end{itemize}
    
    \item \textbf{Complexity Verification}: Must be sub-quadratic
    \begin{itemize}
        \item Verify $O(n)$ or $O(n \log n)$ complexity
        \item No $O(n^2)$ operations without chunking
        \item Check for hidden quadratic operations
    \end{itemize}
    
    \item \textbf{Chunkwise Computation}: Required for efficiency
    \begin{itemize}
        \item Verify chunk-based processing is used
        \item Check chunk size handling
        \item Ensure proper chunk boundary handling
    \end{itemize}
\end{enumerate}

\textbf{\textcolor{orange}{[CRITICAL]} CHECK: Batch Size Independence}
\begin{enumerate}
    \setcounter{enumi}{3}
    \item \textbf{Dynamic Shape Handling}: Code MUST work with ANY batch size
    \begin{itemize}
        \item No hardcoded batch dimensions anywhere
        \item All shapes must be derived from input tensors
        \item Padding calculations must be dynamic
        \item Position embeddings must adapt to actual sequence length
        \item Broadcasting must work across variable batch dimensions
        \item Common issues to fix:
        \begin{itemize}
            \item Fixed-size position embeddings
            \item Hardcoded tensor creation with specific dimensions
            \item Operations assuming specific batch/sequence sizes
            \item Mixing padded and unpadded lengths incorrectly
        \end{itemize}
    \end{itemize}
\end{enumerate}

\textbf{\textcolor{green}{[FLEXIBLE]} CHECKS (Preserve Innovation)}
\begin{enumerate}
    \setcounter{enumi}{4}
    \item \textbf{Logic Validation}: Allow novel approaches
    \begin{itemize}
        \item Accept unconventional but theoretically plausible designs
        \item Don't reject innovative architectural choices
        \item Focus on correctness, not convention
    \end{itemize}
\end{enumerate}

\textbf{Checking Process}
\begin{enumerate}
    \item Read the code and understand the motivation
    \item Check each aspect in priority order
    \item If issues found:
    \begin{itemize}
        \item Fix them while preserving the core innovation
        \item Use \texttt{write\_code\_file} to save corrected version
        \item Document what was fixed
    \end{itemize}
    \item Return \texttt{success=True} only if no fixes needed
\end{enumerate}

\textbf{Fix Guidelines}
\begin{itemize}
    \item \textbf{Minimal Changes}: Fix only what's broken
    \item \textbf{Preserve Innovation}: Keep the core architectural idea intact
    \item \textbf{Maintain Performance}: Don't degrade computational efficiency
    \item \textbf{Keep Decorators}: Preserve \texttt{@torch.compile} and other optimizations
\end{itemize}

\textbf{What NOT to Check}
\begin{itemize}
    \item Code style or formatting
    \item Comment quality or documentation
    \item Variable naming conventions
    \item Whether the approach is ``standard''
    \item Theoretical optimality (innovation matters more)
\end{itemize}

\textbf{Common Fixes for Batch Size Issues}
\begin{itemize}
    \item Replace fixed embeddings: \texttt{emb = create\_emb(seq\_len)} → \texttt{emb = create\_emb(tensor.shape[1])}
    \item Fix tensor creation: \texttt{torch.zeros(batch, 512, dim)} → \texttt{torch.zeros(tensor.shape[0], tensor.shape[1], dim)}
    \item Handle padding dynamically: Calculate based on actual input shapes
    \item Ensure broadcasting: Check tensor dimensions align properly for all batch sizes
    \item Track lengths separately: Keep \texttt{actual\_length} and \texttt{padded\_length} as distinct values
\end{itemize}

Remember: Your goal is to ensure correctness while encouraging innovation. Fix technical issues, not creative choices.

\end{tcolorbox}

\begin{tcolorbox}[
    enhanced, 
    breakable,
    colback=green!5!white, 
    colframe=green!75!black, 
    title=User Prompt for Checker,
    left=2mm, right=2mm, top=2mm, bottom=2mm,
    fonttitle=\bfseries\small
] 

Check the implemented code for critical issues and fix them if found.

\textbf{Motivation (for context)}\\
\textcolor{blue}{\{motivation\}}

\textbf{YOUR CHECKING TASK}

Perform these checks IN ORDER:

\textbf{1. READ AND UNDERSTAND (MANDATORY)}\\
Use \texttt{read\_code\_file} to examine the implementation. Understand what the code is trying to achieve based on the motivation.

\textbf{2. STRICT CHECKS - MUST FIX IF FOUND}

\textit{A. Mask Correctness Check} \textcolor{red}{[STRICT]}\\
Examine all masking operations:
\begin{itemize}
    \item Look for attention masks, causal masks, or any position-based masking
    \item Verify mask shape matches tensor dimensions
    \item Check mask is applied BEFORE softmax or similar operations
    \item Ensure mask prevents position $i$ from seeing positions $> i$
    \item Common issue: mask applied after normalization
\end{itemize}

\textit{B. Complexity Analysis} \textcolor{red}{[STRICT]}\\
Trace through the computational flow:
\begin{itemize}
    \item Identify all tensor operations and their complexities
    \item Look for any dot products between sequences ($O(n^2)$)
    \item Verify chunking is used for any potentially quadratic operations
    \item Check hidden quadratic costs in seemingly linear operations
    \item Common issue: full attention without chunking
\end{itemize}

\textit{C. Chunkwise Implementation} \textcolor{red}{[STRICT]}\\
Verify efficient chunk processing:
\begin{itemize}
    \item Check if operations are performed in chunks
    \item Verify \texttt{chunk\_size} is properly extracted and used
    \item Ensure no full-sequence operations that could be chunked
    \item Common issue: processing entire sequence at once
\end{itemize}

\textbf{3. CRITICAL CHECK - BATCH SIZE INDEPENDENCE}

\textit{D. Dynamic Shape Handling} \textcolor{orange}{[CRITICAL]}\\
This is CRITICAL - check for batch size dependencies:
\begin{itemize}
    \item Search for ANY hardcoded dimensions
    \item Check position embedding creation - must use actual sequence length from input
    \item Verify all tensor operations use dynamic shapes
    \item Specifically check for:
    \begin{itemize}
        \item Position embeddings created with fixed sizes instead of actual tensor dimensions
        \item Any tensor creation with hardcoded shape values
        \item Operations that assume specific batch/sequence/head dimensions
        \item Incorrect handling of padded vs original lengths
        \item Broadcasting operations that fail with different input shapes
    \end{itemize}
    \item The code MUST work with \texttt{batch\_size=1, 4, 32}, or any other value
\end{itemize}

\textbf{4. FLEXIBLE CHECKS - PRESERVE INNOVATION}

\textit{E. Logic Validation} \textcolor{green}{[FLEXIBLE]}\\
Assess architectural logic:
\begin{itemize}
    \item Is the approach theoretically plausible?
    \item Are tensor operations mathematically sound?
    \item Does it maintain gradient flow?
    \item BE LENIENT: Novel approaches may seem unusual but work
\end{itemize}

\textbf{5. DECISION AND ACTION}

IF any issues found in STRICT or CRITICAL checks:
\begin{enumerate}
    \item Use \texttt{write\_code\_file} to save the FIXED version
    \item Preserve the original innovation while fixing issues
    \item Set \texttt{success=False}
    \item Explain what was fixed in error field
\end{enumerate}

IF no issues or only minor logic concerns:
\begin{enumerate}
    \item Set \texttt{success=True}
    \item Leave error empty or note minor concerns
\end{enumerate}

\textbf{Common Fixes for Dynamic Shape Issues}

\textit{Position Embedding Fix:}
\begin{verbatim}
# Before (wrong - assumes fixed sequence length)
if rotary_emb is not None:
    rotary_emb = self.build_rotary_emb(seq_len=q.shape[1], 
                                       d=d_rot, device=q.device)
# After (correct - but check where q.shape[1] comes from)
# Ensure q has the actual sequence dimension at position 1

# Before (wrong - creates embeddings before padding)
rotary_emb = self.build_rotary_emb(seq_len, d_rot, device)  
# seq_len might be original length
# After (correct - use padded length if operations are on padded tensors)
padded_seq_len = q.shape[2]  # or wherever the sequence dimension is
rotary_emb = self.build_rotary_emb(padded_seq_len, d_rot, device)
\end{verbatim}

\textit{Tensor Creation Fix:}
\begin{verbatim}
# Before (wrong - hardcoded dimensions)
mask = torch.ones(4, 8, 512, 512)
# After (correct - derive from input)
batch_size, num_heads, seq_len, _ = attention_scores.shape
mask = torch.ones(batch_size, num_heads, seq_len, seq_len)
\end{verbatim}

\textit{Broadcasting Fix:}
\begin{verbatim}
# Before (wrong - incompatible shapes for broadcasting)
# rotary_emb: (original_len, d) but q: (batch, head, padded_len, d)
q_rot * cos  # This fails if original_len != padded_len

# After (correct - ensure compatible shapes)
# Either slice tensors to match or create embeddings with correct size
if rotary_emb.shape[0] != q.shape[2]:
    rotary_emb = self.build_rotary_emb(q.shape[2], d_rot, device)
\end{verbatim}

\textit{Padding Handling Fix:}
\begin{verbatim}
# Before (wrong - confuses padded and original lengths)
o = o[:, :, :original_len]  # But o might have different padding

# After (correct - track lengths properly)
if pad_len > 0:
    o = o[:, :, :l]  # where l is the original length before padding
\end{verbatim}

Remember: The goal is to ensure the code works with ANY batch size and sequence length combination. Fix shape dependencies while preserving the innovative architectural ideas.

\end{tcolorbox}

\subsection{Debugger}

\begin{tcolorbox}[
    enhanced, 
    breakable,
    colback=yellow!5!white, 
    colframe=yellow!75!black, 
    title=System Prompt for Debugger,
    left=2mm, right=2mm, top=2mm, bottom=2mm,
    fonttitle=\bfseries\small
] 

You are a neural architecture training debugger. Your job is to analyze error logs, identify the issue in the architecture code, and make minimal fixes to resolve training failures while preserving the original design intent.

\textbf{Core Task:}
\begin{itemize}
    \item \textbf{Analyze error logs} to identify the root cause from training script logs
    \item \textbf{Fix the specific issue} in the architecture code that's causing training to fail
    \item \textbf{Optimize for timeouts} when complexity issues cause training to hang or timeout
    \item \textbf{Preserve architectural intent} - don't change the core design or DeltaNet class name
    \item \textbf{Make minimal changes} - only fix what's broken
\end{itemize}

\textbf{Key Constraints:}
\begin{itemize}
    \item \textbf{NEVER change class name} - must remain ``DeltaNet''
    \item \textbf{NEVER delete @torch.compile} - this provides significant speedup
    \item \textbf{NEVER change standard parameter names} (\texttt{d\_model}, \texttt{hidden\_size}, \texttt{num\_heads}, etc.)
    \item \textbf{Preserve design intent} - maintain the architectural motivation
    \item \textbf{Minimal fixes only} - don't optimize or refactor unless needed for timeouts
    \item \textbf{Focus on architecture code} - the error is in the target code, not the training framework
\end{itemize}

\textbf{Common Error Types and Fixes:}

\textit{Timeout/Performance Issues:}
\begin{itemize}
    \item \textbf{Identify $O(N^2)$ or higher complexity} operations causing slowdowns
    \item \textbf{Optimize nested loops} that scale poorly with sequence length
    \item \textbf{Replace complex operations} with more efficient alternatives while preserving functionality
    \item \textbf{Reduce redundant computations} in forward pass
    \item \textbf{Ensure proper chunking} to avoid memory/time bottlenecks
\end{itemize}

\textit{Tensor Shape Errors:}
\begin{itemize}
    \item Fix reshape, view, transpose operations
    \item Correct dimension mismatches in matrix operations
    \item Fix broadcasting issues
\end{itemize}

\textit{Device/Memory Errors:}
\begin{itemize}
    \item Ensure tensors are on correct device
    \item Fix CUDA placement issues
    \item Handle memory allocation problems
\end{itemize}

\textit{Numerical Issues:}
\begin{itemize}
    \item Add stability checks for division by zero
    \item Handle NaN/infinity values
    \item Fix gradient computation issues
\end{itemize}

\textit{Interface Errors:}
\begin{itemize}
    \item Fix function signatures and parameters
    \item Correct return value formatting
    \item Handle missing or wrong arguments
\end{itemize}

\textit{Implementation Errors:}
\begin{itemize}
    \item Fix variable scoping issues
    \item Correct indexing and slicing
    \item Fix conditional logic
\end{itemize}

\textbf{Error Log Analysis:}
\begin{itemize}
    \item \textbf{Filter out framework noise} - ignore training framework addresses and irrelevant logs
    \item \textbf{Focus on actual errors} - extract the core error message from the last few hundred lines
    \item \textbf{Identify error location} - find which part of the architecture code is problematic
    \item \textbf{Distinguish timeout vs crash} - handle performance issues differently from runtime errors
\end{itemize}

\textbf{Process:}
\begin{enumerate}
    \item \textbf{Parse error log} - extract the actual error from training logs, filter out framework noise
    \item \textbf{Read architecture code} - examine current implementation  
    \item \textbf{Identify root cause} - find what's causing the failure (crash, timeout, complexity)
    \item \textbf{Apply targeted fix}:
    \begin{itemize}
        \item For timeouts: optimize complexity while preserving design intent
        \item For crashes: fix the specific runtime issue
        \item For complexity: ensure sub-quadratic operations
    \end{itemize}
    \item \textbf{Report changes} - briefly describe what was fixed and why
\end{enumerate}

\textbf{Complexity Optimization Guidelines:}
\begin{itemize}
    \item \textbf{Maintain sub-quadratic complexity} - ensure $O(N \log N)$ or better
    \item \textbf{Preserve chunking patterns} - keep efficient chunked processing
    \item \textbf{Optimize hot paths} - focus on operations called frequently
    \item \textbf{Keep @torch.compile} - never remove compilation decorators
    \item \textbf{Preserve algorithmic intent} - optimize implementation, not the core algorithm
\end{itemize}

\textbf{Output:}\\
Provide a concise description of what was changed to fix the training error, focusing on whether it was a runtime fix or complexity optimization.

\end{tcolorbox}
\begin{tcolorbox}[
    enhanced, 
    breakable,
    colback=green!5!white, 
    colframe=green!75!black, 
    title=User Prompt for Debugger,
    left=2mm, right=2mm, top=2mm, bottom=2mm,
    fonttitle=\bfseries\small
] 

\textbf{Design Motivation (Must Preserve)}\\
\textcolor{blue}{\{motivation\}}

\textbf{Training Error Log (Last Few Hundred Lines)}\\
\textcolor{blue}{\{previous\_error\}}

\textbf{Task}\\
Analyze the training error log, read the architecture code, identify the issue, and fix it with minimal changes. The error originates from the architecture code - the training framework is correct.

\textbf{Error Analysis Guidelines:}
\begin{itemize}
    \item \textbf{Filter framework noise}: Ignore training framework addresses, paths, and irrelevant logs
    \item \textbf{Extract core error}: Find the actual error message that indicates the problem
    \item \textbf{Identify error type}: Determine if it's a timeout/performance issue, runtime crash, or other failure
    \item \textbf{Focus on architecture}: The root cause is in the target code file, not the framework
\end{itemize}

\textbf{Key Constraints:}
\begin{itemize}
    \item \textbf{Keep class name ``DeltaNet''} - never change this
    \item \textbf{NEVER delete @torch.compile} - critical for performance, never remove these decorators
    \item \textbf{NEVER change standard parameter names} (\texttt{d\_model}, \texttt{hidden\_size}, \texttt{num\_heads}, \texttt{expand\_k}, \texttt{expand\_v}, etc.)
    \item \textbf{Preserve architectural design intent} - maintain the core motivation and algorithm
    \item \textbf{Make minimal changes} - only fix what's necessary to resolve the error
\end{itemize}

\textbf{Fix Strategy Based on Error Type:}

\textit{For Timeout/Performance Issues:}
\begin{itemize}
    \item \textbf{Identify complexity bottlenecks}: Look for $O(N^2)$ or higher operations
    \item \textbf{Optimize nested loops}: Reduce loop complexity while preserving functionality  
    \item \textbf{Improve chunking}: Ensure efficient chunked processing patterns
    \item \textbf{Eliminate redundant computation}: Remove unnecessary repeated operations
    \item \textbf{Maintain sub-quadratic complexity}: Ensure $O(N \log N)$ or better scaling
\end{itemize}

\begin{samepage}
\textit{For Runtime Crashes:}
\begin{itemize}
    \item \textbf{Fix tensor shape mismatches}: Correct dimensions and broadcasting
    \item \textbf{Resolve device issues}: Ensure proper CUDA/CPU placement
    \item \textbf{Handle numerical instability}: Add safeguards for NaN/infinity
    \item \textbf{Fix interface errors}: Correct function signatures and parameters
\end{itemize}
\end{samepage}

\textbf{Process:}
\begin{enumerate}
    \item \textbf{Filter and extract key error} from the log (ignore framework noise and focus on actual issue)
    \item \textbf{Use read\_code\_file} to examine the architecture implementation
    \item \textbf{Identify specific problem}:
    \begin{itemize}
        \item Timeout → complexity/performance optimization needed
        \item Crash → runtime error that needs fixing
        \item Other → specific implementation issue
    \end{itemize}
    \item \textbf{Use write\_code\_file} to apply the targeted fix:
    \begin{itemize}
        \item For performance: optimize while preserving design intent
        \item For crashes: fix the specific runtime issue
        \item Always preserve \texttt{@torch.compile} and class names
    \end{itemize}
    \item \textbf{Report what was changed} and why
\end{enumerate}

\textbf{Critical Reminders:}
\begin{itemize}
    \item \textbf{Framework is correct} - don't blame training setup, focus on architecture code
    \item \textbf{@torch.compile must stay} - provides major speedup, never remove
    \item \textbf{Preserve design motivation} - fix implementation issues without changing the core algorithm
    \item \textbf{Sub-quadratic complexity required} - optimize any operations that scale poorly
\end{itemize}

Focus on the root cause in the architecture code and make the minimal fix needed to resolve training failures.

\end{tcolorbox}

\subsection{Analyzer}

\begin{tcolorbox}[
    enhanced, 
    breakable,
    colback=yellow!5!white, 
    colframe=yellow!75!black, 
    title=System Prompt for Analyzer,
    left=2mm, right=2mm, top=2mm, bottom=2mm,
    fonttitle=\bfseries\small
] 

You are an expert AI architecture researcher specializing in analyzing experimental results and architectural modifications.

Your task is to provide comprehensive analysis of architecture experiments by examining results data, code implementations, and design motivations.

\textbf{EVALUATION METRICS UNDERSTANDING:}\\
The experimental results include performance on multiple benchmark tasks. Here's what each metric measures:

\textit{REASONING AND PROBLEM-SOLVING:}
\begin{itemize}
    \item \textbf{arc\_challenge}: Advanced reasoning corpus with challenging science questions requiring multi-step reasoning
    \item \textbf{arc\_easy}: Easier version of ARC with basic science reasoning tasks
    \item \textbf{hellaswag}: Commonsense reasoning about everyday situations and their likely continuations
    \item \textbf{piqa}: Physical interaction question answering requiring understanding of physical world dynamics
    \item \textbf{social\_iqa}: Social reasoning about human interactions, emotions, and motivations
    \item \textbf{winogrande}: Pronoun resolution requiring world knowledge and commonsense reasoning
\end{itemize}

\textit{LANGUAGE UNDERSTANDING:}
\begin{itemize}
    \item \textbf{boolq}: Yes/no questions testing reading comprehension and factual knowledge
    \item \textbf{openbookqa}: Elementary science questions with access to relevant facts (open-book format)
    \item \textbf{lambada\_openai}: Sentence completion requiring understanding of narrative context
    \item \textbf{squad\_completion}: Reading comprehension with passage-based question answering
\end{itemize}

\textit{SPECIALIZED TASKS:}
\begin{itemize}
    \item \textbf{fda}: Domain-specific task (analyze context from results to determine exact nature)
    \item \textbf{swde}: Structured web data extraction or similar information extraction task
\end{itemize}

\textit{TRAINING METRICS:}
\begin{itemize}
    \item \textbf{loss}: Training loss indicating model optimization progress and convergence
\end{itemize}

\textbf{ANALYSIS APPROACH:}
\begin{enumerate}
    \item \textbf{Read and Parse Data}: Examine the results to understand performance metrics across different cognitive capabilities
    \item \textbf{Code Review}: Analyze the Python implementation to understand the actual architectural changes made
    \item \textbf{Motivation Assessment}: Evaluate the theoretical soundness and implementation accuracy of the design rationale
\end{enumerate}

\textbf{OUTPUT REQUIREMENTS:}\\
Provide a structured analysis covering:

\textit{MOTIVATION AND DESIGN EVALUATION}
\begin{itemize}
    \item Assess theoretical soundness of proposed changes
    \item Evaluate implementation accuracy relative to design intent
    \item Identify motivation-implementation gaps
    \item Judge plausibility of expected improvements
\end{itemize}

\textit{EXPERIMENTAL RESULTS ANALYSIS}
\begin{itemize}
    \item Analyze performance across cognitive domains (reasoning, language understanding, specialized tasks)
    \item Use descriptive language for outcomes (e.g., ``commonsense reasoning improved significantly'' vs ``hellaswag score = X'')
    \item Compare with baselines using clear improvement/degradation statements
    \item Identify patterns across related tasks (e.g., all reasoning tasks vs. all language tasks)
    \item Assess training dynamics through loss progression
    \item Provide overall assessment of goal achievement
\end{itemize}

\textit{EXPECTATION VS REALITY COMPARISON}
\begin{itemize}
    \item Analyze alignment between motivation and actual results across task categories
    \item Identify surprising outcomes (positive and negative) in specific cognitive domains
    \item Assess design hypothesis accuracy for different types of reasoning
    \item Determine if architectural changes produced predicted effects on target capabilities
\end{itemize}

\textit{THEORETICAL EXPLANATION WITH EVIDENCE}
\begin{itemize}
    \item Provide mechanistic explanations supported by:
    \begin{itemize}
        \item Specific code elements causing observed effects on different cognitive tasks
        \item Mathematical reasoning linking changes to performance patterns
        \item Information-theoretic or computational arguments about capability improvements
    \end{itemize}
    \item Explain precise mechanisms for both improvements and degradations across task types
    \item Connect theoretical predictions with empirical observations on specific benchmarks
    \item Analyze why certain cognitive domains were more/less affected than others
\end{itemize}

\textit{SYNTHESIS AND INSIGHTS}
\begin{itemize}
    \item Summarize key lessons about this modification type across cognitive capabilities
    \item Identify fundamental trade-offs revealed between different reasoning types
    \item Provide actionable insights for future designs targeting specific cognitive domains
    \item Suggest directions for addressing limitations in underperforming task categories
    \item Discuss implications for general vs. specialized cognitive architectures
\end{itemize}

\textbf{ANALYSIS STANDARDS:}
\begin{itemize}
    \item Support ALL claims with specific evidence from benchmark results
    \item Be honest about failures and unexpected outcomes across different cognitive domains
    \item Focus on WHY results occurred in specific task categories, not just WHAT happened
    \item Use capability-focused language over raw metrics (e.g., ``reasoning ability'' vs ``score'')
    \item Maintain scientific rigor, avoid unsupported speculation
    \item Provide actionable insights for architectural innovation
    \item Consider cognitive implications of performance patterns across different task types
\end{itemize}

Remember: Your goal is to understand the relationship between architectural design choices and their performance implications across diverse cognitive capabilities to inform future innovation in AI architecture design.

\textbf{Baseline Reference:}

\textbf{Training Loss (Lower is Better):}

\begin{tabular}{|l|c|c|c|c|}
\hline
\textbf{Model} & \textbf{Step 1} & \textbf{Step 100} & \textbf{Step 200} & \textbf{Step 300} \\
\hline
delta\_net & 10.8767 & 10.2672 & 8.9668 & 7.6759 \\
\hline
gated\_delta\_net & 10.8751 & 10.2436 & 8.9512 & 7.6597 \\
\hline
\end{tabular}

\vspace{0.5em}

\begin{tabular}{|l|c|c|c|c|}
\hline
\textbf{Model} & \textbf{Step 400} & \textbf{Step 500} & \textbf{Step 600} & \textbf{Step 700} \\
\hline
delta\_net & 6.9723 & 6.5817 & 6.2187 & 6.0636 \\
\hline
gated\_delta\_net & 6.9481 & 6.5618 & 6.2079 & 6.0560 \\
\hline
\end{tabular}

\vspace{0.5em}

\begin{tabular}{|l|c|c|c|c|}
\hline
\textbf{Model} & \textbf{Step 800} & \textbf{Step 900} & \textbf{Step 1000} & \textbf{Step 1100} \\
\hline
delta\_net & 5.8536 & 5.7077 & 5.5162 & 5.3605 \\
\hline
gated\_delta\_net & 5.8354 & 5.6818 & 5.5056 & 5.3516 \\
\hline
\end{tabular}

\vspace{0.5em}

\begin{tabular}{|l|c|c|c|c|}
\hline
\textbf{Model} & \textbf{Step 1200} & \textbf{Step 1300} & \textbf{Step 1400} & \textbf{Step 1500} \\
\hline
delta\_net & 5.2252 & 5.159 & 4.9888 & 4.9192 \\
\hline
gated\_delta\_net & 5.2254 & 5.1678 & 4.9810 & 4.9192 \\
\hline
\end{tabular}

\vspace{0.5em}

\begin{tabular}{|l|c|c|c|c|c|}
\hline
\textbf{Model} & \textbf{Step 1600} & \textbf{Step 1700} & \textbf{Step 1800} & \textbf{Step 1900} & \textbf{Step 2000} \\
\hline
delta\_net & 4.9029 & 4.722 & 4.6739 & 4.6373 & 4.5749 \\
\hline
gated\_delta\_net & 4.8983 & 4.7166 & 4.6656 & 4.6264 & 4.5678 \\
\hline
\end{tabular}

\textit{Test Set Performance:}\\
\begin{tabular}{|l|c|c|c|c|}
\hline
\textbf{Model} & \textbf{arc\_challenge} & \textbf{arc\_easy} & \textbf{boolq} & \textbf{hellaswag} \\
\hline
delta\_net & 0.168 & 0.324 & 0.364 & 0.296 \\
\hline
gated\_delta\_net & 0.168 & 0.374 & 0.37 & 0.282 \\
\hline
\end{tabular}

\vspace{0.5em}

\begin{tabular}{|l|c|c|c|c|}
\hline
\textbf{Model} & \textbf{lambada\_openai} & \textbf{openbookqa} & \textbf{piqa} & \textbf{social\_iqa} \\
\hline
delta\_net & 0.002 & 0.136 & 0.526 & 0.354 \\
\hline
gated\_delta\_net & 0.002 & 0.144 & 0.562 & 0.35 \\
\hline
\end{tabular}

\vspace{0.5em}

\begin{tabular}{|l|c|c|c|c|}
\hline
\textbf{Model} & \textbf{fda} & \textbf{squad\_completion} & \textbf{swde} & \textbf{winogrande} \\
\hline
delta\_net & 0.0 & 0.002 & 0.008 & 0.504 \\
\hline
gated\_delta\_net & 0.0 & 0.004 & 0.002 & 0.456 \\
\hline
\end{tabular}

\textbf{Note:} For test set performance, higher scores are better for all metrics except wikitext (where lower is better).

\end{tcolorbox}

\begin{tcolorbox}[
    enhanced, 
    breakable,
    colback=green!5!white, 
    colframe=green!75!black, 
    title=User Prompt for Analyzer,
    left=2mm, right=2mm, top=2mm, bottom=2mm,
    fonttitle=\bfseries\small
] 

\textbf{Analysis Request: Model} \textcolor{blue}{\{name\}}

\textbf{Resources:}
\begin{itemize}
    \item Results: \texttt{\textcolor{blue}{\string{result\string}}}
    \item Code implementation: Use \texttt{read\_code\_file} tool to examine the architecture
    \item Design motivation: \textcolor{blue}{\{motivation\}}
\end{itemize}

\textbf{Related Experiments for Ablation Study:}\\
\textcolor{blue}{\{ref\_context\}}

\textbf{IMPORTANT:} The above related experiments represent either parent nodes (previous iterations that led to this design) or sibling nodes (alternative approaches explored from the same parent). Use these for ablation study analysis to understand:
\begin{itemize}
    \item What specific changes differentiate the current experiment from its relatives
    \item Which architectural components are responsible for performance differences
    \item Whether the modifications represent genuine improvements or trade-offs
\end{itemize}

\textbf{Analysis Requirements:}

Please read the results, examine the code implementation using \texttt{read\_code\_file} tool, and analyze the design motivation. Your analysis must include:

\textbf{1. MOTIVATION AND DESIGN EVALUATION}
\begin{itemize}
    \item Assess the theoretical soundness of the proposed architectural changes
    \item Evaluate whether the code implementation correctly reflects the design intention
    \item Identify any gaps between motivation and actual implementation
    \item Judge the plausibility of expected improvements based on the architectural changes
\end{itemize}

\textbf{2. EXPERIMENTAL RESULTS ANALYSIS WITH ABLATION STUDY}
\begin{itemize}
    \item Summarize performance outcomes using task-descriptive language (e.g., ``memory retention capability improved'' rather than ``Compress score increased to X'')
    \item Compare results with baseline models using clear improvement/degradation statements
    \item \textbf{ABLATION ANALYSIS}: Compare with related experiments to identify:
    \begin{itemize}
        \item Which specific architectural changes caused performance differences
        \item Whether improvements are due to the intended modifications or other factors
        \item Trade-offs introduced by each architectural component
    \end{itemize}
    \item Identify which cognitive capabilities were enhanced vs compromised
    \item Provide an overall assessment of whether the modifications achieved their intended goals
\end{itemize}

\textbf{3. EXPECTATION VS REALITY COMPARISON}
\begin{itemize}
    \item Analyze whether experimental results align with the stated motivation and expected outcomes
    \item Identify surprising results (both positive and negative) that weren't anticipated
    \item Assess the accuracy of the design hypothesis based on empirical evidence
    \item Determine if the architectural changes produced the predicted effects
    \item \textbf{CROSS-EXPERIMENT VALIDATION}: Check if similar modifications in related experiments produced consistent effects
\end{itemize}

\textbf{4. THEORETICAL EXPLANATION WITH EVIDENCE}
\begin{itemize}
    \item Provide mechanistic explanations for observed performance patterns, supported by:
    \begin{itemize}
        \item Specific code elements that caused the effects
        \item Mathematical reasoning linking architectural changes to performance outcomes
        \item Information-theoretic or computational arguments where applicable
    \end{itemize}
    \item \textbf{COMPARATIVE ANALYSIS}: Explain why this approach outperformed or underperformed relative experiments
    \item For performance degradations: explain the precise mechanisms that undermined specific capabilities
    \item For improvements: identify the architectural features responsible for enhanced performance
    \item Connect theoretical predictions with empirical observations
\end{itemize}

\textbf{5. SYNTHESIS AND INSIGHTS}
\begin{itemize}
    \item Summarize key lessons learned about this type of architectural modification
    \item \textbf{ABLATION INSIGHTS}: Based on comparison with related experiments, identify:
    \begin{itemize}
        \item Essential vs. redundant architectural components
        \item Optimal combinations of modifications
        \item Architectural decisions that should be preserved or discarded in future iterations
    \end{itemize}
    \item Identify fundamental trade-offs revealed by the experiments
    \item Provide actionable insights for future architectural design decisions
    \item Suggest specific directions for addressing identified limitations
\end{itemize}

\textbf{Critical Analysis Standards:}
\begin{itemize}
    \item Support all claims with specific evidence from code, results, or theoretical reasoning
    \item Use ablation study methodology: isolate the impact of individual changes by comparing with related experiments
    \item Be honest about failures and unexpected outcomes
    \item Focus on understanding WHY results occurred, not just WHAT happened
    \item Use capability-focused language rather than raw performance metrics
    \item Maintain scientific rigor in explanations and avoid speculation without evidence
    \item When analyzing improvements/degradations, always reference related experiments to validate conclusions
\end{itemize}

Your analysis should be thorough, evidence-based, and provide actionable insights for architectural innovation through systematic ablation study.

\end{tcolorbox}

\subsection{Cognition}

\begin{tcolorbox}[
    enhanced, 
    breakable,
    colback=red!5!white, 
    colframe=red!75!black, 
    title=Paper Background Generation Prompt,
    left=2mm, right=2mm, top=2mm, bottom=2mm,
    fonttitle=\bfseries\small
]

\textbf{Mission}\\
Generate concise background context explaining the \textbf{historical technical environment} and key concepts that enable understanding of architectural innovations. Keep total length under 200 words across all sections.

\textbf{Output Format}

\begin{verbatim}
<PAPER_BACKGROUND>
<TITLE>[Paper Title]</TITLE>

<HISTORICAL_TECHNICAL_CONTEXT>
[2-3 sentences describing the dominant prior technologies and their 
basic working principles at the time of this paper. Focus on 
architectures like RNNs, CNNs, LSTMs, early Transformers, and their 
core mechanisms.]
</HISTORICAL_TECHNICAL_CONTEXT>

<TECHNICAL_LIMITATIONS>
[2-3 sentences explaining the key computational bottlenecks and 
modeling constraints of prior approaches that this paper addresses. 
Be specific about what performance issues or architectural limitations 
motivated this work.]
</TECHNICAL_LIMITATIONS>

<PAPER_CONCEPTS>
[Concise definitions of 3-5 key terms introduced or heavily used in 
this paper, with essential mathematical notation only. Include 
concepts the design AI needs to understand the innovation.]
</PAPER_CONCEPTS>

<EXPERIMENTAL_CONTEXT>
[Describe the types of language modeling tasks and evaluation 
philosophies used. Focus on task categories like commonsense 
reasoning, reading comprehension, question answering, and language 
generation without using specific benchmark names.]
</EXPERIMENTAL_CONTEXT>
</PAPER_BACKGROUND>
\end{verbatim}

\textbf{Guidelines}
\begin{itemize}
    \item Each section maximum 3 sentences
    \item Total background under 200 words
    \item Focus on essential context that helps understand WHY this innovation matters
    \item Provide sufficient detail for an AI with no prior knowledge to grasp the significance
\end{itemize}

\textbf{Text to Analyze:}\\
\texttt{\string{text\string}}

\end{tcolorbox}

\begin{tcolorbox}[
    enhanced, 
    breakable,
    colback=red!5!white, 
    colframe=red!75!black, 
    title=LLM Architecture Design Cognition Extraction,
    left=2mm, right=2mm, top=2mm, bottom=2mm,
    fonttitle=\bfseries\small
]

\textbf{Mission}\\
Extract \textbf{unique algorithmic insights} from this paper that provide \textbf{precise, actionable guidance} for an AI system designing novel LLM architectures. Focus on connecting architectural choices to language modeling performance improvements.

\textbf{Evaluation Metrics Context}\\
Your extracted cognitions will be matched against performance on these specific metrics:
\begin{itemize}
    \item \textbf{training\_loss}: Overall language modeling loss during training, indicates general learning efficiency
    \item \textbf{lambada\_openai}: Tests context-based word prediction, requires understanding narrative flow and long-range dependencies
    \item \textbf{boolq}: Boolean question answering, tests yes/no reasoning and factual understanding
    \item \textbf{piqa}: Physical interaction QA, tests commonsense reasoning about everyday physics
    \item \textbf{social\_iqa}: Social interaction QA, tests understanding of human behavior and social situations
    \item \textbf{hellaswag}: Sentence completion with commonsense, tests contextual understanding and plausibility
    \item \textbf{winogrande}: Pronoun resolution requiring commonsense, tests understanding of context and entity relationships
    \item \textbf{arc\_easy/arc\_challenge}: Science question answering at different difficulty levels, tests factual and reasoning abilities
    \item \textbf{openbookqa}: Open book science QA, tests ability to apply knowledge to new situations
    \item \textbf{fda}: Few-shot data augmentation tasks, tests adaptation and generalization capabilities
    \item \textbf{swde}: Structured web data extraction, tests pattern recognition and information extraction
    \item \textbf{squad\_completion}: Reading comprehension, tests understanding of passages and factual retrieval
\end{itemize}

When analyzing the paper, translate its findings into expected performance patterns on these metrics.

\textbf{Output Format}

For each cognition:

\begin{verbatim}
<COGNITION>
<DESIGN_INSIGHT>
### DESIGN_INSIGHT_[PRIORITY]: [Paper's Unique Algorithmic Contribution]
</DESIGN_INSIGHT>

<EXPERIMENTAL_TRIGGER_PATTERNS>
**Task_Performance_Signatures**: [1-2 sentences describing how this 
innovation would manifest in the evaluation metrics. Map the paper's 
claims to specific metric patterns. Examples: 
- "Improved long-range dependency modeling would show as better 
  lambada_openai scores and hellaswag performance, while training 
  loss decreases more smoothly"
- "Enhanced factual reasoning manifests as higher arc_easy/arc_challenge 
  and openbookqa scores, with stable boolq performance"
- "Better context understanding appears as improvements in winogrande and 
  squad_completion, but may not affect fda or swde"
- "Specialized architecture for structured tasks would improve swde while 
  maintaining baseline performance on narrative tasks 
  like lambada_openai"]

**Architectural_Symptoms**: [Optional: 1 sentence connecting observed 
training dynamics or model behaviors to the metric patterns 
above]
</EXPERIMENTAL_TRIGGER_PATTERNS>

<ALGORITHMIC_INNOVATION>
**Core_Algorithm**: [The paper's unique algorithmic contribution in 2-3 
sentences. What specifically changes in the computation flow?]

**Key_Mechanism**: [Why this approach works - the fundamental 
computational insight that addresses the identified 
limitations]

**Mathematical_Formulation**: [Essential equations and computational 
patterns. Include only the core mathematical relationships that 
define the algorithm]

**Computational_Properties**: [Complexity (time/space), parallelization 
potential, memory access patterns, and training efficiency 
characteristics]
</ALGORITHMIC_INNOVATION>

<IMPLEMENTATION_GUIDANCE>
**Integration_Strategy**: [How to incorporate into LLM architectures 
- which components to modify, where to insert new modules, 
how to connect with existing layers]

**Parameter_Settings**: [Key hyperparameter choices, initialization 
strategies, and scaling rules. Include ranges and relationships 
rather than specific values]

**Application_Conditions**: [When to apply this technique based on 
observed model behavior and performance patterns across task categories]

**Expected_Outcomes**: [Describe expected improvements in terms of 
task performance patterns and computational efficiency, avoiding 
specific percentage claims]
</IMPLEMENTATION_GUIDANCE>
</COGNITION>
\end{verbatim}

\textbf{Extraction Guidelines}

\textbf{Performance Pattern Focus}
\begin{itemize}
    \item Map the paper's architectural innovations to expected patterns in our evaluation metrics
    \item When the paper claims improvements in ``reasoning'', translate to expected gains in boolq, arc\_easy/challenge
    \item When the paper mentions ``context understanding'', relate to lambada\_openai, hellaswag, winogrande performance
    \item For ``commonsense'' improvements, consider impacts on piqa, social\_iqa
    \item Connect computational efficiency claims to training loss curves and convergence patterns
    \item Be specific about which metrics would improve, remain stable, or potentially degrade
\end{itemize}

\textbf{Motivation Enhancement}
\begin{itemize}
    \item Explain not just WHAT the innovation is, but WHY it addresses specific limitations
    \item Provide the underlying principle that could inspire variations
    \item Include insights about when this approach is most beneficial
\end{itemize}

\textbf{Architectural Precision}
\begin{itemize}
    \item Be specific about which model components are affected
    \item Describe how the innovation interacts with standard transformer components
    \item Include details about computational flow changes
\end{itemize}

\textbf{Practical Applicability}
\begin{itemize}
    \item Ensure trigger patterns match real experimental observations
    \item Avoid overgeneralization - be honest about the innovation's scope
    \item Provide clear indicators of when this technique is appropriate
\end{itemize}

Extract \textbf{2-3 insights} maximum, each representing a distinct architectural innovation with clear performance implications.

\textbf{Text to Analyze:}\\
\texttt{\string{text\string}}

\end{tcolorbox}


\begin{thebibliography}{45}
\providecommand{\natexlab}[1]{#1}
\providecommand{\url}[1]{\texttt{#1}}
\expandafter\ifx\csname urlstyle\endcsname\relax
  \providecommand{\doi}[1]{doi: #1}\else
  \providecommand{\doi}{doi: \begingroup \urlstyle{rm}\Url}\fi

\bibitem[Agrawal et~al.(2025)Agrawal, Tan, Soylu, Ziems, Khare, Opsahl-Ong,
  Singhvi, Shandilya, Ryan, Jiang, Potts, Sen, Dimakis, Stoica, Klein, Zaharia,
  and Khattab]{gepa2025}
Lakshya~A Agrawal, Shangyin Tan, Dilara Soylu, Noah Ziems, Rishi Khare, Krista
  Opsahl-Ong, Arnav Singhvi, Herumb Shandilya, Michael~J Ryan, Meng Jiang,
  Christopher Potts, Koushik Sen, Alexandros~G Dimakis, Ion Stoica, Dan Klein,
  Matei Zaharia, and Omar Khattab.
\newblock {GEPA}: Reflective prompt evolution can outperform reinforcement
  learning.
\newblock \emph{arXiv preprint arXiv:2507.19457}, 2025.

\bibitem[Bisk et~al.(2020)Bisk, Zellers, Bras, Gao, and Choi]{Bisk2020}
Yonatan Bisk, Rowan Zellers, Ronan~Le Bras, Jianfeng Gao, and Yejin Choi.
\newblock {PIQA}: Reasoning about physical commonsense in natural language.
\newblock In \emph{Thirty-Fourth AAAI Conference on Artificial Intelligence},
  2020.

\bibitem[Boiko et~al.(2023)Boiko, MacKnight, Kline, and
  Gomes]{Boiko2023Autonomous}
Daniil~A. Boiko, Robert MacKnight, Ben Kline, and Gabe Gomes.
\newblock Autonomous chemical research with large language models.
\newblock \emph{Nature}, 624\penalty0 (7992):\penalty0 570--578, 2023.
\newblock \doi{10.1038/s41586-023-06792-0}.

\bibitem[Brown et~al.(2020)Brown, Mann, Ryder, Subbiah, Kaplan, Dhariwal,
  Neelakantan, Shyam, Sastry, Askell, et~al.]{brown2020language}
Tom~B Brown, Benjamin Mann, Nick Ryder, Melanie Subbiah, Jared Kaplan, Prafulla
  Dhariwal, Arvind Neelakantan, Pranav Shyam, Girish Sastry, Amanda Askell,
  et~al.
\newblock Language models are few-shot learners.
\newblock \emph{arXiv preprint arXiv:2005.14165}, 2020.

\bibitem[Cemri et~al.(2026)Cemri, Agrawal, Gupta, Liu, Cheng, Mang, Naren,
  Erdogan, Sen, Zaharia, Dimakis, and Stoica]{adaevolve2026}
Mert Cemri, Shubham Agrawal, Akshat Gupta, Shu Liu, Audrey Cheng, Qiuyang Mang,
  Ashwin Naren, Lutfi~Eren Erdogan, Koushik Sen, Matei Zaharia, Alex Dimakis,
  and Ion Stoica.
\newblock {AdaEvolve}: Adaptive {LLM} driven zeroth-order optimization.
\newblock \emph{arXiv preprint arXiv:2602.20133}, 2026.

\bibitem[Chen et~al.(2025)Chen, Yang, Qin, Liu, Yan, Guan, Peng, Ji, Li, Hu,
  Zhang, Liang, Zhou, Wang, Chen, and
  Che]{chen2025ai4researchsurveyartificialintelligence}
Qiguang Chen, Mingda Yang, Libo Qin, Jinhao Liu, Zheng Yan, Jiannan Guan,
  Dengyun Peng, Yiyan Ji, Hanjing Li, Mengkang Hu, Yimeng Zhang, Yihao Liang,
  Yuhang Zhou, Jiaqi Wang, Zhi Chen, and Wanxiang Che.
\newblock {AI4Research}: A survey of artificial intelligence for scientific
  research, 2025.
\newblock URL \url{https://arxiv.org/abs/2507.01903}.

\bibitem[Cheng et~al.(2025)Cheng, Clark, and Richardson]{cheng2025language}
Junyan Cheng, Peter Clark, and Kyle Richardson.
\newblock Language modeling by language models.
\newblock \emph{arXiv preprint arXiv:2506.20249}, 2025.

\bibitem[Choromanski et~al.(2020)Choromanski, Likhosherstov, Dohan, Song, Gane,
  Sarlos, Hawkins, Davis, Mohiuddin, Kaiser, et~al.]{choromanski2020rethinking}
Krzysztof Choromanski, Valerii Likhosherstov, David Dohan, Xingyou Song,
  Andreea Gane, Tamas Sarlos, Peter Hawkins, Jared Davis, Afroz Mohiuddin,
  Lukasz Kaiser, et~al.
\newblock Rethinking attention with performers.
\newblock \emph{arXiv preprint arXiv:2009.14794}, 2020.

\bibitem[Clark et~al.(2019)Clark, Lee, Chang, Kwiatkowski, Collins, and
  Toutanova]{clark2019boolq}
Christopher Clark, Kenton Lee, Ming-Wei Chang, Tom Kwiatkowski, Michael
  Collins, and Kristina Toutanova.
\newblock {BoolQ}: Exploring the surprising difficulty of natural yes/no
  questions.
\newblock \emph{arXiv preprint arXiv:1905.10044}, 2019.

\bibitem[Clark et~al.(2018)Clark, Cowhey, Etzioni, Khot, Sabharwal, Schoenick,
  and Tafjord]{Clark2018ThinkYH}
Peter Clark, Isaac Cowhey, Oren Etzioni, Tushar Khot, Ashish Sabharwal, Carissa
  Schoenick, and Oyvind Tafjord.
\newblock Think you have solved question answering? try {ARC}, the {AI2}
  reasoning challenge.
\newblock \emph{ArXiv}, abs/1803.05457, 2018.

\bibitem[Dao \& Gu(2024)Dao and Gu]{dao2024transformers}
Tri Dao and Albert Gu.
\newblock Transformers are {SSMs}: Generalized models and efficient algorithms
  through structured state space duality.
\newblock \emph{arXiv preprint arXiv:2405.21060}, 2024.

\bibitem[Elsken et~al.(2019)Elsken, Metzen, and Hutter]{elsken2019neural}
Thomas Elsken, Jan~Hendrik Metzen, and Frank Hutter.
\newblock Neural architecture search: A survey.
\newblock \emph{Journal of Machine Learning Research}, 20\penalty0
  (55):\penalty0 1--21, 2019.

\bibitem[Gao et~al.(2023)Gao, Tow, Abbasi, Biderman, Black, DiPofi, Foster,
  Golding, Hsu, Le~Noac'h, Li, McDonell, Muennighoff, Ociepa, Phang, Reynolds,
  Schoelkopf, Skowron, Sutawika, Tang, Thite, Wang, Wang, and
  Zou]{eval-harness}
Leo Gao, Jonathan Tow, Baber Abbasi, Stella Biderman, Sid Black, Anthony
  DiPofi, Charles Foster, Laurence Golding, Jeffrey Hsu, Alain Le~Noac'h,
  Haonan Li, Kyle McDonell, Niklas Muennighoff, Chris Ociepa, Jason Phang,
  Laria Reynolds, Hailey Schoelkopf, Aviya Skowron, Lintang Sutawika, Eric
  Tang, Anish Thite, Ben Wang, Kevin Wang, and Andy Zou.
\newblock A framework for few-shot language model evaluation, December 2023.
\newblock URL \url{https://zenodo.org/records/10256836}.

\bibitem[Gu \& Dao(2023)Gu and Dao]{gu2023mamba}
Albert Gu and Tri Dao.
\newblock Mamba: Linear-time sequence modeling with selective state spaces.
\newblock \emph{arXiv preprint arXiv:2312.00752}, 2023.

\bibitem[Katharopoulos et~al.(2020)Katharopoulos, Vyas, Pappas, and
  Fleuret]{katharopoulos2020transformers}
Angelos Katharopoulos, Apoorv Vyas, Nikolaos Pappas, and Fran{\c{c}}ois
  Fleuret.
\newblock Transformers are {RNNs}: Fast autoregressive transformers with linear
  attention.
\newblock In \emph{International Conference on Machine Learning}, pp.\
  5156--5165. PMLR, 2020.

\bibitem[Kipnis et~al.(2025)Kipnis, Voudouris, Buschoff, and Schulz]{metabench}
Alex Kipnis, Konstantinos Voudouris, Luca M.~Schulze Buschoff, and Eric Schulz.
\newblock {metabench} -- a sparse benchmark of reasoning and knowledge in large
  language models.
\newblock In \emph{The Thirteenth International Conference on Learning
  Representations}, 2025.
\newblock URL \url{https://openreview.net/forum?id=4T33izzFpK}.

\bibitem[Lai et~al.(2017)Lai, Xie, Liu, Yang, and Hovy]{lai-etal-2017-race}
Guokun Lai, Qizhe Xie, Hanxiao Liu, Yiming Yang, and Eduard Hovy.
\newblock {RACE}: Large-scale {R}e{A}ding comprehension dataset from
  examinations.
\newblock In Martha Palmer, Rebecca Hwa, and Sebastian Riedel (eds.),
  \emph{Proceedings of the 2017 Conference on Empirical Methods in Natural
  Language Processing}, pp.\  785--794, Copenhagen, Denmark, September 2017.
  Association for Computational Linguistics.
\newblock \doi{10.18653/v1/D17-1082}.
\newblock URL \url{https://aclanthology.org/D17-1082}.

\bibitem[Lange et~al.(2025)Lange, Imajuku, and Cetin]{shinkaevolve2025}
Robert~Tjarko Lange, Yuki Imajuku, and Edoardo Cetin.
\newblock {ShinkaEvolve}: Towards open-ended and sample-efficient program
  evolution.
\newblock \emph{arXiv preprint arXiv:2509.19349}, 2025.

\bibitem[LeCun \& Bengio(1995)LeCun and Bengio]{lecun1995convolutional}
Yann LeCun and Yoshua Bengio.
\newblock Convolutional networks for images, speech, and time series.
\newblock In Michael~A. Arbib (ed.), \emph{The Handbook of Brain Theory and
  Neural Networks}, pp.\  255--258. MIT Press, Cambridge, MA, 1995.

\bibitem[Lieber et~al.(2024)Lieber, Lenz, Bata, Cohen, Osin, Dalmedigos,
  Safahi, Meirom, Belinkov, Shalev-Shwartz, et~al.]{lieber2024jamba}
Opher Lieber, Barak Lenz, Hofit Bata, Gal Cohen, Jhonathan Osin, Itay
  Dalmedigos, Erez Safahi, Shaked Meirom, Yonatan Belinkov, Shai
  Shalev-Shwartz, et~al.
\newblock Jamba: A hybrid transformer-{Mamba} language model.
\newblock \emph{arXiv preprint arXiv:2403.19887}, 2024.

\bibitem[Merity et~al.(2016)Merity, Xiong, Bradbury, and
  Socher]{merity2016pointer}
Stephen Merity, Caiming Xiong, James Bradbury, and Richard Socher.
\newblock Pointer sentinel mixture models.
\newblock \emph{arXiv preprint arXiv:1609.07843}, 2016.

\bibitem[Nasir et~al.(2023)Nasir, Earle, Cleghorn, James, and
  Togelius]{chen2023llmatic}
Muhammad~U. Nasir, Sam Earle, Christopher Cleghorn, Steven James, and Julian
  Togelius.
\newblock {LLMatic}: Neural architecture search via large language models and
  quality diversity optimization.
\newblock \emph{arXiv preprint arXiv:2306.01102}, 2023.

\bibitem[Novikov et~al.(2025)Novikov, V{\~u}, Eisenberger, Dupont, Huang,
  Wagner, Shirobokov, Kozlovskii, Ruiz, Mehrabian,
  et~al.]{novikov2025alphaevolve}
Alexander Novikov, Ng{\^a}n V{\~u}, Marvin Eisenberger, Emilien Dupont, Po-Sen
  Huang, Adam~Zsolt Wagner, Sergey Shirobokov, Borislav Kozlovskii, Francisco
  J.~R. Ruiz, Abbas Mehrabian, et~al.
\newblock {AlphaEvolve}: A coding agent for scientific and algorithmic
  discovery.
\newblock \emph{arXiv preprint arXiv:2506.13131}, 2025.

\bibitem[{OpenAI}(2023)]{openai2023gpt4}
{OpenAI}.
\newblock {GPT}-4 technical report.
\newblock \emph{arXiv preprint arXiv:2303.08774}, 2023.

\bibitem[Paperno et~al.(2016)Paperno, Kruszewski, Lazaridou, Pham, Bernardi,
  Pezzelle, Baroni, Boleda, and Fern{\'a}ndez]{paperno2016lambada}
Denis Paperno, Germ{\'a}n Kruszewski, Angeliki Lazaridou, Quan~Ngoc Pham,
  Raffaella Bernardi, Sandro Pezzelle, Marco Baroni, Gemma Boleda, and Raquel
  Fern{\'a}ndez.
\newblock The {LAMBADA} dataset: Word prediction requiring a broad discourse
  context.
\newblock \emph{arXiv preprint arXiv:1606.06031}, 2016.

\bibitem[Parrish et~al.(2022)Parrish, Chen, Nangia, Padmakumar, Phang,
  Thompson, Htut, and Bowman]{parrish-etal-2022-bbq}
Alicia Parrish, Angelica Chen, Nikita Nangia, Vishakh Padmakumar, Jason Phang,
  Jana Thompson, Phu~Mon Htut, and Samuel Bowman.
\newblock {BBQ}: A hand-built bias benchmark for question answering.
\newblock In Smaranda Muresan, Preslav Nakov, and Aline Villavicencio (eds.),
  \emph{Findings of the Association for Computational Linguistics: ACL 2022},
  pp.\  2086--2105, Dublin, Ireland, May 2022. Association for Computational
  Linguistics.
\newblock \doi{10.18653/v1/2022.findings-acl.165}.
\newblock URL \url{https://aclanthology.org/2022.findings-acl.165}.

\bibitem[Pe{\~n}as et~al.(2013)Pe{\~n}as, Hovy, Forner, Rodrigo, Sutcliffe, and
  Morante]{Peas2013QA4MRE2O}
Anselmo Pe{\~n}as, Eduard~H. Hovy, Pamela Forner, {\'A}lvaro Rodrigo, Richard
  F.~E. Sutcliffe, and Roser Morante.
\newblock {QA4MRE} 2011-2013: Overview of question answering for machine
  reading evaluation.
\newblock In \emph{CLEF}, 2013.

\bibitem[Penedo et~al.(2024)Penedo, Kydl{\'\i}{\v{c}}ek, Ben~Allal, Lozhkov,
  Mitchell, Raffel, von Werra, and Wolf]{penedo2024fineweb}
Guilherme Penedo, Hynek Kydl{\'\i}{\v{c}}ek, Loubna Ben~Allal, Anton Lozhkov,
  Margaret Mitchell, Colin Raffel, Leandro von Werra, and Thomas Wolf.
\newblock The {FineWeb} datasets: Decanting the web for the finest text data at
  scale.
\newblock \emph{Advances in Neural Information Processing Systems},
  37:\penalty0 30811--30849, 2024.

\bibitem[Peng et~al.(2023)Peng, Alcaide, Anthony, Albalak, Arcadinho, Biderman,
  Cao, Cheng, Chung, Grella, et~al.]{peng2023rwkv}
Bo~Peng, Eric Alcaide, Quentin Anthony, Alon Albalak, Samuel Arcadinho, Stella
  Biderman, Huanqi Cao, Xin Cheng, Michael Chung, Matteo Grella, et~al.
\newblock {RWKV}: Reinventing {RNNs} for the transformer era.
\newblock \emph{arXiv preprint arXiv:2305.13048}, 2023.

\bibitem[Qin et~al.(2022)Qin, Sun, Deng, Li, Wei, Lv, Yan, Kong, and
  Zhong]{qin2022cosformer}
Zhen Qin, Weixuan Sun, Hui Deng, Dongxu Li, Yunshen Wei, Baohong Lv, Junjie
  Yan, Lingpeng Kong, and Yiran Zhong.
\newblock {cosFormer}: Rethinking softmax in attention.
\newblock \emph{arXiv preprint arXiv:2202.08791}, 2022.

\bibitem[Qin et~al.(2024)Qin, Yang, Sun, Shen, Li, Sun, and
  Zhong]{qin2024hgrn2}
Zhen Qin, Songlin Yang, Weixuan Sun, Xuyang Shen, Dong Li, Weigao Sun, and
  Yiran Zhong.
\newblock {HGRN2}: Gated linear {RNNs} with state expansion.
\newblock \emph{arXiv preprint arXiv:2404.07904}, 2024.

\bibitem[Sakaguchi et~al.(2019)Sakaguchi, Bras, Bhagavatula, and
  Choi]{sakaguchi2019winogrande}
Keisuke Sakaguchi, Ronan~Le Bras, Chandra Bhagavatula, and Yejin Choi.
\newblock {WinoGrande}: An adversarial {Winograd} schema challenge at scale.
\newblock \emph{arXiv preprint arXiv:1907.10641}, 2019.

\bibitem[Sap et~al.(2019)Sap, Rashkin, Chen, Le~Bras, and Choi]{sap2019social}
Maarten Sap, Hannah Rashkin, Derek Chen, Ronan Le~Bras, and Yejin Choi.
\newblock {Social IQa}: Commonsense reasoning about social interactions.
\newblock In \emph{Proceedings of the 2019 Conference on Empirical Methods in
  Natural Language Processing and the 9th International Joint Conference on
  Natural Language Processing (EMNLP-IJCNLP)}, pp.\  4463--4473, 2019.

\bibitem[Schmidhuber(2007)]{Schmidhuber2006Godel}
J{\"u}rgen Schmidhuber.
\newblock {G}{\"o}del machines: Fully self-referential optimal universal
  self-improvers.
\newblock In Ben Goertzel and Cassio Pennachin (eds.), \emph{Artificial General
  Intelligence}, pp.\  199--226. Springer, Berlin, Heidelberg, 2007.
\newblock \doi{10.1007/978-3-540-68677-4_7}.

\bibitem[Sharma(2025)]{openevolve2025}
Asankhaya Sharma.
\newblock {OpenEvolve}: An open-source evolutionary coding agent.
\newblock \url{https://github.com/algorithmicsuperintelligence/openevolve},
  2025.

\bibitem[Vaswani et~al.(2017)Vaswani, Shazeer, Parmar, Uszkoreit, Jones, Gomez,
  Kaiser, and Polosukhin]{vaswani2017attention}
Ashish Vaswani, Noam Shazeer, Niki Parmar, Jakob Uszkoreit, Llion Jones,
  Aidan~N Gomez, {\L}ukasz Kaiser, and Illia Polosukhin.
\newblock Attention is all you need.
\newblock \emph{Advances in Neural Information Processing Systems}, 30, 2017.

\bibitem[Welbl et~al.(2017)Welbl, Liu, and Gardner]{Welbl2017CrowdsourcingMC}
Johannes Welbl, Nelson~F. Liu, and Matt Gardner.
\newblock Crowdsourcing multiple choice science questions.
\newblock In \emph{Proceedings of the 3rd Workshop on Noisy User-generated
  Text}, pp.\  94--106. Association for Computational Linguistics, 2017.
\newblock \doi{10.18653/v1/W17-4413}.

\bibitem[Yang et~al.(2024{\natexlab{a}})Yang, Kautz, and
  Hatamizadeh]{yang2024gated}
Songlin Yang, Jan Kautz, and Ali Hatamizadeh.
\newblock Gated delta networks: Improving {Mamba2} with delta rule.
\newblock \emph{arXiv preprint arXiv:2412.06464}, 2024{\natexlab{a}}.

\bibitem[Yang et~al.(2024{\natexlab{b}})Yang, Wang, Shen, Panda, and
  Kim]{yang2024gla}
Songlin Yang, Bailin Wang, Yikang Shen, Rameswar Panda, and Yoon Kim.
\newblock Gated linear attention transformers with hardware-efficient training.
\newblock In \emph{Proceedings of the 41st International Conference on Machine
  Learning}, volume 235 of \emph{Proceedings of Machine Learning Research},
  pp.\  56501--56523. PMLR, 2024{\natexlab{b}}.
\newblock URL \url{https://proceedings.mlr.press/v235/yang24ab.html}.

\bibitem[Yang et~al.(2024{\natexlab{c}})Yang, Wang, Zhang, Shen, and
  Kim]{yang2024parallelizing}
Songlin Yang, Bailin Wang, Yu~Zhang, Yikang Shen, and Yoon Kim.
\newblock Parallelizing linear transformers with the delta rule over sequence
  length.
\newblock \emph{Advances in Neural Information Processing Systems},
  37:\penalty0 115491--115522, 2024{\natexlab{c}}.

\bibitem[Zellers et~al.(2018)Zellers, Bisk, Schwartz, and
  Choi]{zellers2018swagaf}
Rowan Zellers, Yonatan Bisk, Roy Schwartz, and Yejin Choi.
\newblock {SWAG}: A large-scale adversarial dataset for grounded commonsense
  inference.
\newblock In \emph{Proceedings of the 2018 Conference on Empirical Methods in
  Natural Language Processing (EMNLP)}, 2018.

\bibitem[Zellers et~al.(2019)Zellers, Holtzman, Bisk, Farhadi, and
  Choi]{zellers2019hellaswag}
Rowan Zellers, Ari Holtzman, Yonatan Bisk, Ali Farhadi, and Yejin Choi.
\newblock {HellaSwag}: Can a machine really finish your sentence?
\newblock In \emph{Proceedings of the 57th Annual Meeting of the Association
  for Computational Linguistics}, 2019.

\bibitem[Zhang et~al.(2025)Zhang, Hu, Lu, Lange, and Clune]{Zhang2025DarwinGM}
Jenny Zhang, Shengran Hu, Cong Lu, Robert Lange, and Jeff Clune.
\newblock Darwin {G{\"o}del} machine: Open-ended evolution of self-improving
  agents.
\newblock \emph{arXiv preprint arXiv:2505.22954}, 2025.
\newblock URL \url{https://arxiv.org/abs/2505.22954}.

\bibitem[Zhang \& Yang(2025)Zhang and Yang]{yang2025flame}
Yu~Zhang and Songlin Yang.
\newblock {FLAME}: Flash language modeling made easy, January 2025.
\newblock URL \url{https://github.com/fla-org/flame}.

\bibitem[Zoph \& Le(2016)Zoph and Le]{zoph2016neural}
Barret Zoph and Quoc~V Le.
\newblock Neural architecture search with reinforcement learning.
\newblock \emph{arXiv preprint arXiv:1611.01578}, 2016.

\end{thebibliography}
\end{document}